\documentclass[conference]{IEEEtran}
\IEEEoverridecommandlockouts

\usepackage{cite}
\usepackage{amsmath,amssymb,amsfonts}
\usepackage{algorithmic}
\usepackage{graphicx}
\usepackage{textcomp}
\usepackage{xcolor}
\def\BibTeX{{\rm B\kern-.05em{\sc i\kern-.025em b}\kern-.08em
    T\kern-.1667em\lower.7ex\hbox{E}\kern-.125emX}}

\usepackage{subfig}
\usepackage{array,multirow,graphicx}
\usepackage{booktabs}
\usepackage{algorithmic}
\usepackage{bbm}
\usepackage{algorithm}
\usepackage{mhchem}

\begin{document}

\title{On the Necessity of Multi-Domain Explanation: \\ An Uncertainty Principle Approach for Deep Time Series Models\\
}


\author{\IEEEauthorblockN{Shahbaz Rezaei}
\IEEEauthorblockA{\textit{University of California}\\
Davis, CA, USA \\
srezaei@ucdavis.edu}
\and
\IEEEauthorblockN{Avishai Halev}
\IEEEauthorblockA{\textit{University of California}\\
Davis, CA, USA \\
avishaihalev@gmail.com}
\and
\IEEEauthorblockN{Xin Liu}
\IEEEauthorblockA{\textit{University of California}\\
Davis, CA, USA \\
xinliu@ucdavis.edu}
\and
}

\maketitle

\begin{abstract}
A prevailing approach to explain time series models is to generate attribution in time domain input. A recent development in time series XAI is the concept of explanation spaces, where any model trained in the time domain can be interpreted with any existing XAI method in alternative domains, such as frequency or time-frequency domain. The prevailing approach is to present XAI attributions either in the time domain or in the domain where the attribution is most sparse. In this paper, we demonstrate that in certain cases, XAI methods can generate attributions that highlight fundamentally different features in the time and frequency domains that are not direct counterparts of one another. This observation suggests that both domains' attributions should be presented to achieve a more comprehensive interpretation. Thus it shows the necessity of multi-domain explanation.

To quantify when such cases arise, we introduce the \textit{uncertainty principle} (UP), originally developed in quantum mechanics and later studied in harmonic analysis and signal processing, to the XAI literature. This principle establishes a lower bound on how much a signal can be simultaneously localized in both the time and frequency domains. By leveraging this concept, we assess whether attributions in the time and frequency domains violate this bound, indicating that they emphasize distinct features. In other words, UP provides a sufficient condition that the time and frequency domain explanations do not match and, hence, should be both presented to the end user. We validate the effectiveness of this approach across various deep learning models, XAI methods, and a wide range of classification and forecasting datasets. The frequent occurrence of UP violations across various datasets and XAI methods highlights the limitations of existing approaches that focus solely on time-domain explanations. This underscores the need for multi-domain explanations as a new paradigm. The key contribution of this work is to use an uncertainty principle to show that explanations in both the time and frequency domains are necessary in many time series applications. The source code will be available upon acceptance.

\end{abstract}


\section{Introduction}
As deep learning models become more widespread, it is increasingly important to understand their functionality and performance in downstream tasks. Explainable artificial intelligence (XAI) is a growing field of research dedicated to interpreting machine learning models. One of the most common approaches in XAI, particularly in computer vision, involves highlighting the parts of the input to which the target model is most sensitive, known as attribution-based methods \cite{sundararajan2017axiomatic}. The same algorithms and approaches that have been developed for computer vision has been adopted to time series data and are among the most commonly used.

\begin{figure}[h]
    Case 1:\\
    \subfloat[Time]{%
        \includegraphics[width=0.45\linewidth, height=0.12\textwidth]{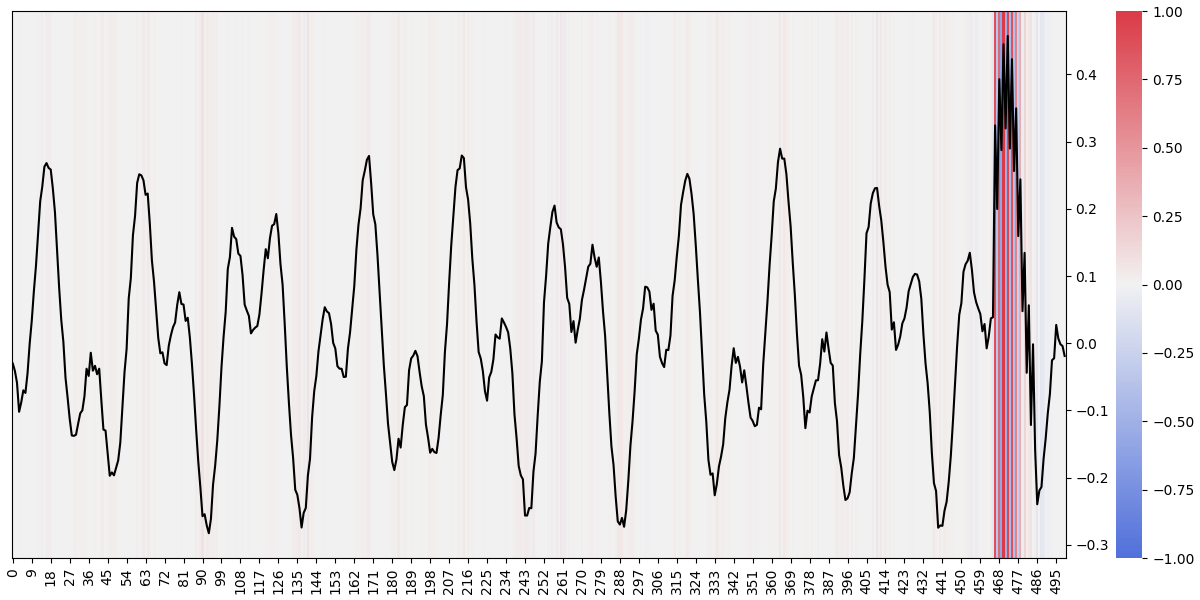}
        }
    \subfloat[Frequency]{
        \includegraphics[width=0.45\linewidth, height=0.12\textwidth]{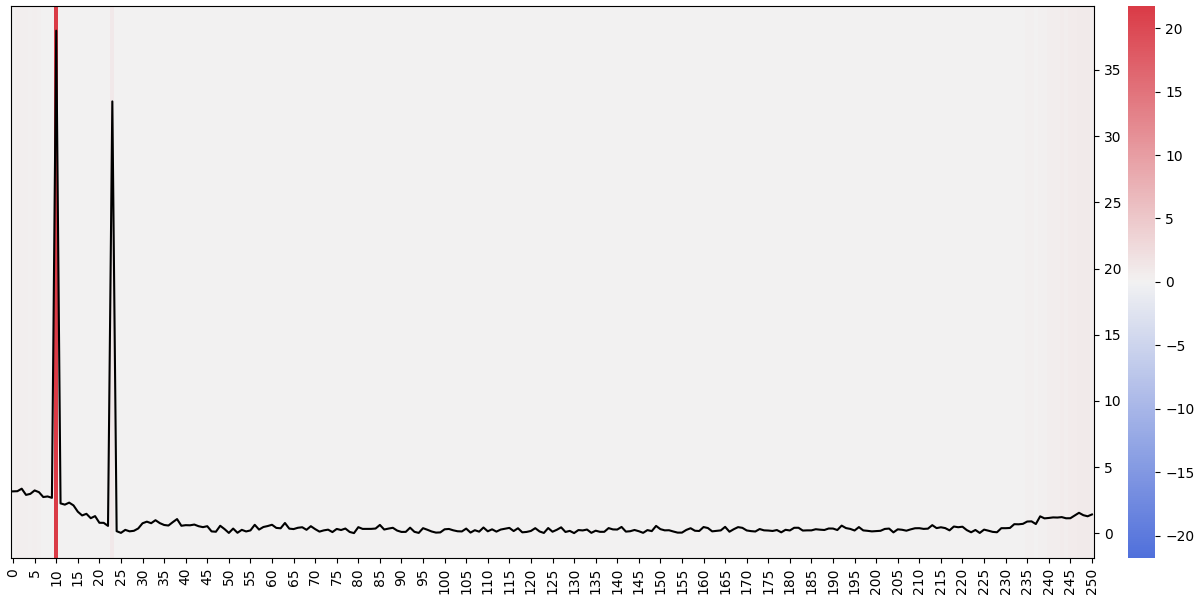}
        }
    \\
    Case 2:\\
    \subfloat[Time]{%
        \includegraphics[width=0.45\linewidth, height=0.12\textwidth]{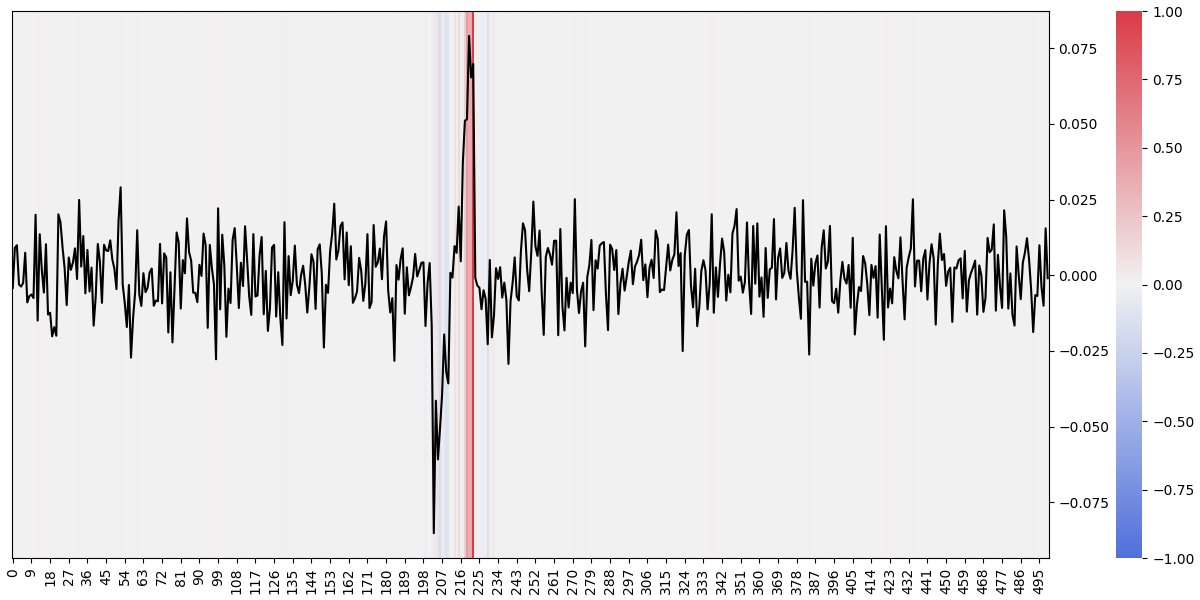}
        }
    \subfloat[Frequency]{
        \includegraphics[width=0.45\linewidth, height=0.12\textwidth]{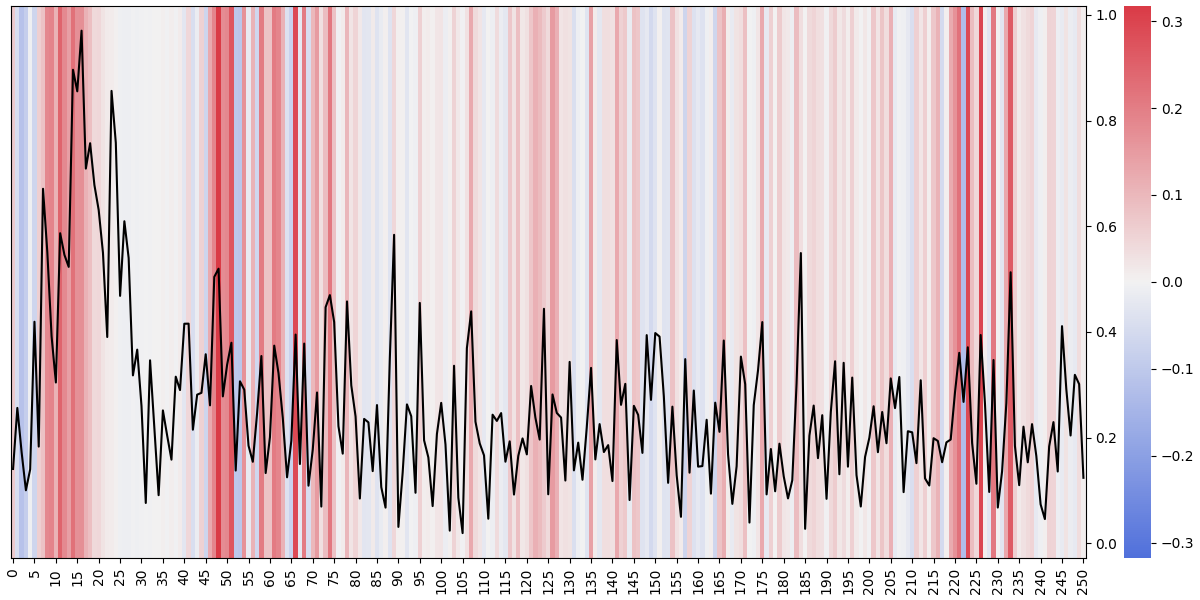}
        }
    \\
    Case 3:\\
    \subfloat[Time]{%
        \includegraphics[width=0.45\linewidth, height=0.12\textwidth]{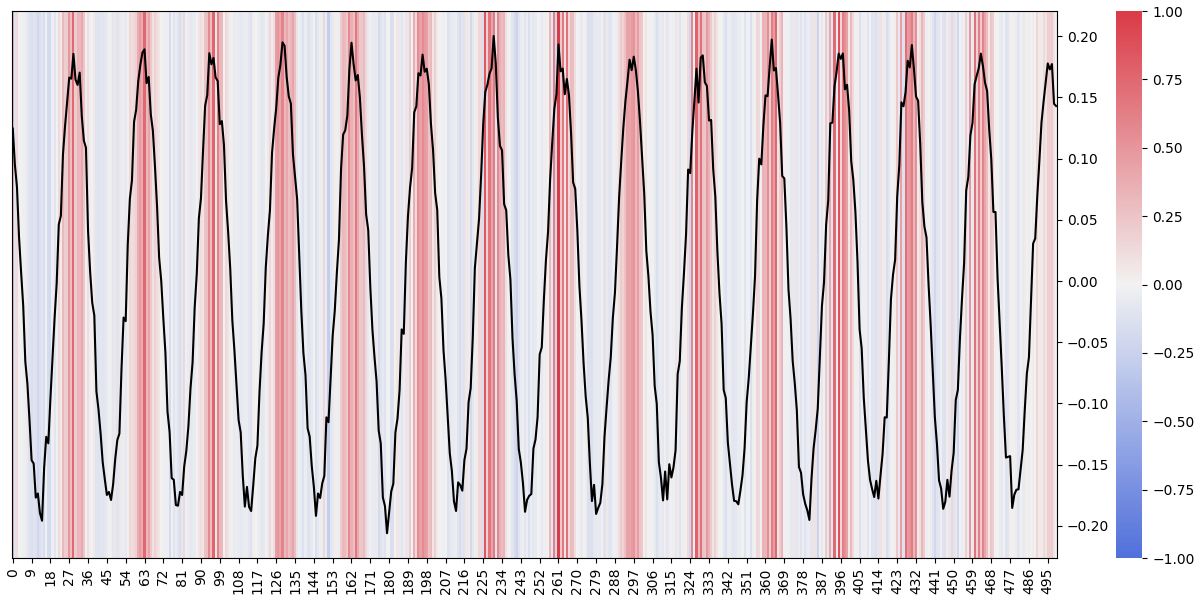}
        }
    \subfloat[Frequency]{
        \includegraphics[width=0.45\linewidth, height=0.12\textwidth]{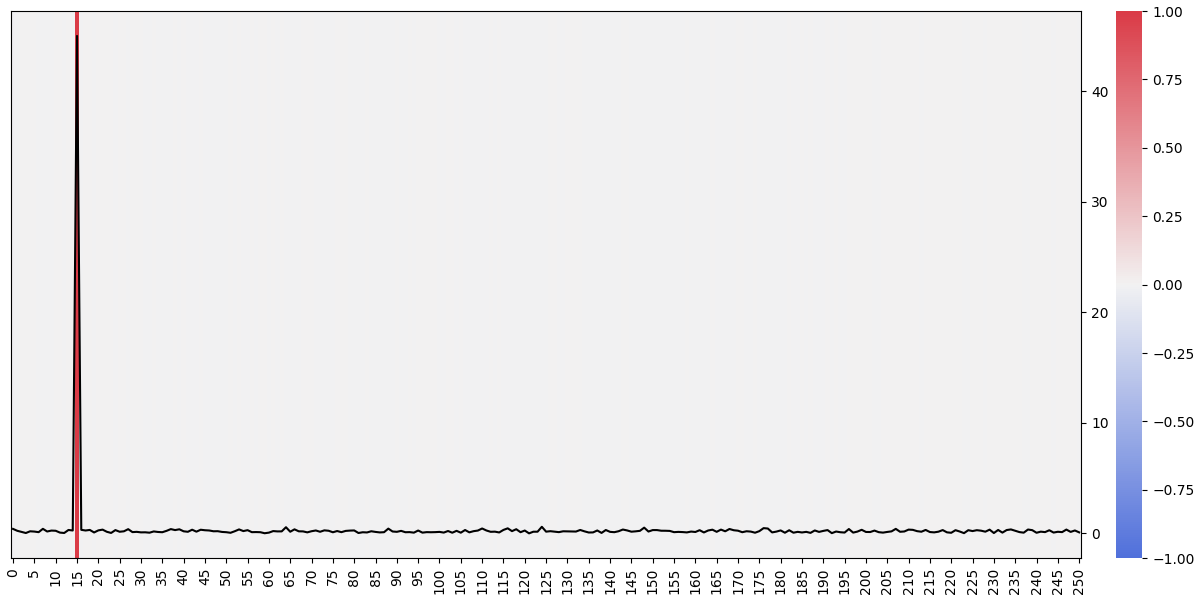}
        }
    \caption{Synthetic dataset containing both/either shapelet features and/or frequency features. The ResNet model is trained on the time domain and explained using DeepLift. The attribution on frequency domain is generated using wrapped method \cite{rezaei2024explanation} using the same sample/model. In case 1, a sample contains both shapelet and dominant frequency feature. DeepLift on time domain highlights the shapelet (a), while the explanation of the same sample in frequency space highlights the frequency feature (b). In case 2 and 3, the sample only contain shapelet and dominant frequency, respectively.
    }
    \label{fig:synthetic}
\end{figure}

Recently, it has been shown that time series contain unique features that require special treatment. For instance, highlighting long-range features such as trends and seasonality is difficult for a user to interpret in the time domain. In \cite{rezaei2024explanation} and \cite{vielhaben2024explainable}, authors have proposed a new way to adopt existing attribution-based methods to highlight what the model has learned in different domains, such as frequency, time/frequency, etc. Such a method is powerful because it does not require the model to be trained on the target domain. The model, trained originally on the time domain, can be explained on another domain using any existing XAI using wrapper method \cite{rezaei2024explanation}. As a result, features, such as seasonality, can be localized\footnote{ In this paper, we use localized and sparse interchangeably.} in the frequency domain and easily understood by the user. Most of the existing literature recommends the domain on which the model is trained on to be presented to the end user while \cite{rezaei2024explanation} recommends the most sparse domain. However, time series data are often rich in information and features, and consequently, deep models may learn a wide range of features from shapelets (i.e. a few consecutive and distinguishable time steps) to frequencies. In this paper, we show that a single explanation space may not be able to capture the full functionality of a deep model.

In this work, we focus on multi-domain explanation, motivated by one key observation: Consider the example presented in Figure \ref{fig:synthetic}. In Figure \ref{fig:synthetic}, we show the attribution, generated by DeepLIFT, of a model trained on the time domain on a synthetic dataset containing shapelet features and/or dominant frequency features (more details in Section \ref{sec:evaluation}). For each case, we use the method outlined in \cite{rezaei2024explanation} to generate attribution on the frequency domain. Case 1 in Figure \ref{fig:synthetic} contains both shapelet and frequency features. While the attribution in the time domain highlights the shapelet feature, the attribution in the frequency domain highlights the dominant frequency. This is not surprising because the frequency feature in the time domain is wide spread. Hence, assuming that the similar value of importance is supposed to be given to both frequency and shapelet features, the importance for frequency feature is distributed to the entire time domain and will not be visibly as highlighted as the shapelet feature. Case 2 (3) illustrates the attribution of a sample that contains only the shapelet (dominant frequency) feature. The attribution is only localized in the time (frequency) domain. 

These examples clearly show that a single explanation space may not be enough to explain a deep model, whether it is time or another domain. A challenging question arises: How can we define a quantitative measure to check if the attribution in the time domain corresponds to the same feature in the frequency domain (case 2 and 3) or not (case 1).

To address this challenge, we bring the theories developed in quantum mechanics and, later on, in signal processing to quantify whether the time and frequency attributions can correspond to the same feature or not. The uncertainty principle (UP), first proposed by Heisenberg in 1927 \cite{heisenberg1927anschaulichen}, states that canonically conjugate variables, such as position and momentum, cannot be both sharply localized \cite{folland1997uncertainty}. This includes a non-zero function and its Fourier transform. Simply put, if a Fourier transform of a function is sharply localized, e.g. containing only a single frequency, the time domain representation is a widespread signal that is not localized. If the time domain representation of a sample is localized, then the Fourier transform is inevitably wide spread and non-localized. In other words, if the attributions in the time and frequency domains are both localized, then the features they represent cannot be the same. 

The uncertainty principle gives a lower bound on how much time and frequency representation can be localized. For a real conjugate pair of a sample and its Fourier transform, the uncertainty principle gives a bound on how much these two can be localized and the bound cannot be violated. In this paper, we adopt the bound to the attributions separately generated in the time and frequency domains. 
When these two attributions highlight different features, as shown in Figure \ref{fig:synthetic}, and because they are not direct Fourier transforms of each other, UP can be violated.
In other words, the UP violation is a sufficient condition that the features represented on time and frequency cannot be the same. Thus, both domains should be presented to the end user. 

The main contribution of this paper is to show that a common practice of a single-domain explanation in times series is insufficient and potentially misleading. We propose using UP as a sufficient condition to show time/frequency explanations are not each other's counterpart. We demonstrate UP violation across various deep learning models and existing XAI methods underscoring the need for multi-domain explanation.


\section{Related Work}

\subsection{Explanation Methods}


Attribution-based methods generate a heat map for a given sample that indicates how much each time step contributes to the model's prediction. Attribution can be generated using gradient, such as GradientSHAP \cite{erion2019learning}, Integrated Gradient \cite{sundararajan2017axiomatic}, DeepLIFT \cite{shrikumar2017learning}, Saliency \cite{simonyan2013deep}, DeepSHAP \cite{lundberg2017unified}, InputXGradient \cite{shrikumar2016not}, 
and KernelSHAP \cite{lundberg2017unified}, or occluding features, such as Feature Occlusion \cite{zeiler2014visualizing}, and feature ablation \cite{suresh2017clinical}. The majority of commonly used attribution methods are first proposed and examined for computer vision tasks, which can be easily adopted for time series application. Recently, a few methods have been proposed to address the specific challenges of time series data, such as TSR \cite{ismail2020benchmarking}, Demux \cite{doddaiah2022class}, TimeX \cite{queen2024encoding}, LEFTIST \cite{guilleme2019agnostic}, MILLET \cite{early2023inherently}, or DFT-LPR \cite{vielhaben2024explainable}.

In this paper, we focus on the most widely used attribution methods. The goal is not to have a comprehensive comparison, but rather to show the UP violation can be utilized for attribution explanation broadly. We show that an explanation of a model trained on the time domain can be different when attribution is generated on the time domain versus the frequency domain. Two algorithms are developed to generate attribution on the frequency domain when a model is originally trained on the time domain: DFT-LPR \cite{vielhaben2024explainable} and the explanation space wrapper \cite{rezaei2024explanation}. DFT-LPR only works for LPR XAI method while the wrapper idea of \cite{rezaei2024explanation} is a general framework that works on any existing attribution methods. In this paper, we use an explanation space wrapper to generate attribution in the frequency domain.

\subsection{Uncertainty Principle}
Uncertainty principle (UP) refers to any of the theorems which quantify the idea that a non-zero function and its Fourier transform cannot be both localized. The first version of the theorem, called the Heisenberg uncertainty principle \cite{heisenberg1927anschaulichen}, was introduced in 1927 to state that the position and momentum of a particle cannot be both measured to an arbitrary precision. This similarity between the Fourier transform and the position/momentum pair arises because the momentum in quantum mechanics is the Fourier operator on the position. This phenomenon has been studied for over a century now with a wide range of applications, including quantum mechanics, signal processing \cite{donoho1989uncertainty}, cryptography \cite{coles2017entropic}, etc. We refer to the comprehensive work of Folland and Sitaram \cite{folland1997uncertainty} for more information on the history of the uncertainty principle. 

There are different ways to measure how much a signal is localized in the time/frequency domain. The original Heisenberg uncertainty principle defines localization in terms of variance and states that the multiplication of variance in time (position) and frequency (momentum) has a lower bound. Many different notions of locality has been studied ever since. For instance, it has been shown that a function, $f$, and its Fourier, $\hat{f}$, cannot both have a compact support \cite{wigderson2021uncertainty}. In another branch of uncertainty principles, it has been shown that $f$ and $\hat{f}$ cannot both have supports of finite measure \cite{benedicks1985fourier}. Some other theorems define localization as a decay of a function at infinity. For instance, Hardy shows that $f$ and $\hat{f}$ cannot be both decay faster than $e^{-x^2}$ \cite{hardy1933theorem}. The other well-studied versions are called logarithmic uncertainty principles which define localization in terms of Shannon entropy \cite{beckner1975inequalities}. Based on its applications in signal processing, a new type of uncertainty principle was proposed in \cite{donoho1989uncertainty} that deals with finite groups rather than functions on $\mathbb{R}$. Although many of the existing uncertainty principles can be potentially adopted for time series applications, we focus on a version proposed in \cite{donoho1989uncertainty} because of its tight bound and the fact that it can easily handle discrete time data.

\section{Method}

\subsection{Uncertainty Principle}

\textbf{Discrete Fourier transform:} Let $x_t$ be a sequence of length $N$ and $x_f$ be the corresponding discrete Fourier transform, as follows:
\begin{equation}
    x_f \equiv \frac{1}{\sqrt{N}} \sum_{t=0}^{N-1}{x_t e^{-2 \pi i w t /N}},     f=0, ..., N-1.
\end{equation}

In this paper, we use the uncertainty principles proposed in \cite{donoho1989uncertainty}. The simplest version of the uncertainty principle is stated in Theorem 1 and Corollary 1.

\textbf{Theorem 1.} Let $N_t$ and $N_f$ count the number of non-zero entries in $x_t$ and $x_f$, then
\begin{equation}
        N_t  N_f \geqq N.
\end{equation}

\textbf{Corollary 1.} 
\begin{equation}
        N_t +  N_f \geqq 2 \sqrt{N}.
\end{equation}

Here, we mainly focus on the uncertainty principle theorems that deal with a sequence $x_t$ and its corresponding Fourier transform $x_f$, rather than continuous functions. In UP theorems, the pair is guaranteed not to be violated. Here, motivated by the example shown in Figure \ref{fig:synthetic} where it is possible for an attribution on time and frequency domain to be both localized, we use the uncertainty principle to see a potential violation. UP violation, therefore, indicates that the features highlighted on time and frequency domains are not the same; otherwise, at least one domain would have not been localized.

To achieve this goal, we assume that the target deep model is trained only on the time domain\footnote{ One can also train the model on frequency domain and perform the same procedure. However, we ignore this case because it is not common practice.}. As a result, we can easily generate an attribution in the time domain using existing XAI methods. This attribution is used as $x_t$ in the uncertainty principle formulation. To generate attribution in the frequency domain, we use the explanation space wrapper \cite{rezaei2024explanation}. 
The wrapper method embeds the model trained on time domain and attaches a head before the model's input that applies inverse FFT. Hence, a time domain sample can be converted to the frequency domain using FFT and fed to the wrapper model. Hence, the attribution is generated on the frequency domain. This generates an attribution on the frequency domain which we use as $x_f$ in the uncertainty principle formalism. Note that here time and frequency attributions are generated independently. In other words, time and frequency attributions are not Fourier counterparts of each others. That is why plugging these values into the uncertainty principle may violate it.

An alternative approach to generate attribution in the frequency domain is to take the Fourier transform of the time domain attribution. 
However, taking the Fourier transform of the attribution in Figure \ref{fig:synthetic} (a) would not reveal the frequency feature shown in Figure \ref{fig:synthetic} (b). The reason why using wrapper idea may reveal different features is simple. Taking the easiest case of the occlusion method of XAI, occluding a single step or a few consecutive steps can reveal a shapelet on the time domain while it may barely change the overall shape of a frequency wave, and, hence the model's output. On the other hand, occluding the frequency feature in the frequency domain can reveal the dominant frequency feature. Hence, if both frequency and shapelet features exist in a sample, XAI methods might be able to localized each in the corresponding domain. However, taking the Fourier transform of the attribution on the time domain cannot highlight the dominant frequency. Therefore, we use the wrapper method and generate attributions on time and frequency domains separately.

Despite the simplicity of the uncertainty principles in Theorem 1 and Corollary 1, it is not a practically useful bound for attribution pairs. The reason is that most XAI methods assign non-zero values to all time steps. For example, in Figure \ref{fig:synthetic}, the portion of the attribution that seems not to be highlighted are not actually exactly zero. Hence, $N_f$ and $N_t$ are always equal to $N$ and the lower-bound is trivial. Consequently, the uncertainty principle will never be violated. For this reason, we use the following theorem from \cite{donoho1989uncertainty} which allows us to remove the effects of the very small non-zero values.

\textbf{Theorem 2:} Let $x_t$ be $\epsilon_t$-concentrated on the index set $T$ and $x_f$ be $\epsilon_f$-concentrated on the index set $F$. Let $N_t$ and $N_f$ denote the number of elements in $T$ and $F$, respectively. Then, 
\begin{equation}
\label{eq:epsilon_based_UP}
    N_t N_f \geqq N (1 - (\epsilon_t + \epsilon_f))^2.
\end{equation}

A sequence, $q=(q_n)$, is $\epsilon$-concentrated on an index set $R$ if most of the energy is contained in $R$, that is 
\begin{equation}
\label{eq:epsilon_concentration}
    (\sum_{n \notin R}{(q_n)^2})^{1/2} \leq \epsilon .
\end{equation}

\subsection{Measuring UP Violation}
We use equation (\ref{eq:epsilon_based_UP}) to measure the UP violation. The equation should hold for every pair of epsilons, $(\epsilon_t, \epsilon_f)$, that matches the criteria of the Theorem. 

To see if UP is violated according to Equation (\ref{eq:epsilon_based_UP}), we can perform a grid search based on $(N_t,N_f)$ pairs.
Because they are discrete natural numbers, we do not need to search for values in between. If we choose the tightest corresponding $(\epsilon_t, \epsilon_f)$ pairs for a given $(N_t,N_f)$ and it does not violate the uncertainty principle, it guarantees that no other epsilon pair would violate the UP because any other pair would decrease the left-hand side of the equation (\ref{eq:epsilon_based_UP}) while the right-hand side remains the same. The tightest bound is achieved by sorting all values and choosing the smallest $N_t$ or $N_f$ values depending on the domain. Hence, the complexity of the algorithm is $O(N^2)$ as shown in Algorithm \ref{algo:UP}. Here, we first sort the values in $(x_t, x_f)$ from the smallest to the largest (line 4 and 5). For a given $N_t$, we choose the first $N_t$ smallest values in $X_t$ to put outside the index set $T$. By doing so, we obtain the smallest $\epsilon_t$ (line 10) for a given $N_t$ since any other set $T'$ of size $N_t$ would yield larger $\epsilon$ in equation (\ref{eq:epsilon_concentration}). We perform a similar operation for $x_f$ and $N_f$. If UP is violated for any pair of $(N_t,N_f)$, the algorithm is terminated, otherwise the UP is not violated.

\renewcommand{\algorithmicrequire}{\textbf{Input:}}

\begin{algorithm}
\begin{algorithmic}[1]
\REQUIRE $x_t$, $x_f$
\STATE $N \gets |x_t|$
\STATE $x_t \gets Abs(Unit\_norm(x_t))$
\STATE $x_f \gets Abs(Unit\_norm(x_f))$
\STATE $t_{steps} \gets Sort(x_t)$
\STATE $f_{steps} \gets Sort(x_f)$
\FOR{$i\gets 1, N$}
\FOR{$j\gets 1, N$}
\STATE $\bar{\epsilon}_t \gets t_{steps}[i]$
\STATE $\bar{\epsilon}_f \gets f_{steps}[j]$
\STATE $N_t \gets |x_t \mathbbm{1}_{\{x_t > \bar{\epsilon}_t\}}|$
\STATE $N_f \gets |x_f \mathbbm{1}_{\{x_f > \bar{\epsilon}_f\}}|$
\STATE $\epsilon_t \gets \lVert x_t \mathbbm{1}_{\{x_t \leq \bar{\epsilon}_t\}} \rVert_2$
\STATE $\epsilon_f \gets \lVert x_f \mathbbm{1}_{\{x_f \leq \bar{\epsilon}_f\}} \rVert_2$
\IF {$\epsilon_t + \epsilon_f< 1$}
\IF {$ N_t N_f < N(1 - (\epsilon_t + \epsilon_f))^2$}
    \STATE return UP\_Violation
\ENDIF 
\ENDIF 
\ENDFOR
\ENDFOR
\caption{UP Violation Algorithm}
\label{algo:UP}
\end{algorithmic}
\end{algorithm}

\section{Evaluation}
\label{sec:evaluation}

\textbf{Experimental Setup: } We train five common deep learning models: fully connected network (FCN), ResNet, InceptionTime \cite{ismail2020inceptiontime}, temporal convolutional network (TCN) \cite{bai2018empirical}, and time series transformer (TST) \cite{zerveas2021transformer}. The FCN has three layers of size (256, 128, 64) with ReLU nonlinearity. All other models are adopted from times series AI package \cite{tsai}. All models are trained on only the time domain using the Adam optimizer except TST for which we use AdamW. For regression tasks, we use the MSE loss. Due to the limited computational capacity, we choose a batch size of 4 for training TST models which explains the suboptimal performance of TST models. The explanations are generated on different domains using the explanation space wrapper idea of \cite{rezaei2024explanation}. We evaluate seven well-known XAI methods: DeepLIFT, GradientSHAP, InputXGradient (I$\times$G), Integrated Gradient (IG), LIME \cite{ribeiro2016should}, Occlusion, and Saliency \cite{simonyan2013deep}. We use the Captum implementation \cite{kokhlikyan2020captum} for XAI methods. The DeepLIFT method is incompatible with TCN. So, we do not report its results. Other details of the experiments are similar to \cite{rezaei2024explanation}.

\begin{table*}[ht]
\scriptsize
  \centering
  \begin{tabular}{lllllllllll}
    \toprule
    Validation Set & Model & Accuracy & Activation & DeepLIFT & GradSHAP & I $\times$ G & IG & Occlusion & Saliency & LIME\\
    \midrule   
    Time/Freq. (class 0) & FCN & 100\% & 3.57 $\pm$ 1.59 & 0.00\% & 0.00\% & 0.00\% & 0.00\% & 0.00\% & 0.00\%  & 100\% \\
    Time/Freq. (class 1) & FCN & 100\% & 3.36 $\pm$ 1.26 & 0.00\% & 2.00\% & 2.00\% & 0.00\% & 2.00\% & 0.00\% & 98.00\% \\
    Time (class 0) & FCN & 90.0\% & 0.80 $\pm$ 0.71 & 0.00\% & 4.00\% & 0.00\% & 2.00\% & 2.00\% & 0.00\% & 98.00\% \\
    Time (class 1) & FCN & 100\% & 1.05 $\pm$ 0.33 & 0.00\% & 0.00\% & 0.00\% & 0.00\% & 4.00\% & 0.00\% & 98.00\% \\
    Freq. (class 0) & FCN & 100\% & 2.95 $\pm$ 0.93 & 0.00\% & 0.00\% & 0.00\% & 0.00\% & 0.00\% & 0.00\% & 98.00\% \\
    Freq. (class 1) & FCN & 100\% & 3.01 $\pm$ 0.64 & 0.00\% & 0.00\% & 0.00\% & 0.00\% & 2.00\% & 0.00\% & 100\% \\
    
    \midrule
    Time/Freq. (class 0) & ResNet & 86\% & 1.71 $\pm$ 1.73 & 76.00\% & 18.00\% & 8.00\% & 40.00\% & 8.00\% & 0.00\% & 98.00\% \\
    Time/Freq. (class 1) & ResNet & 100\% & 2.51 $\pm$ 0.64 & 0.00\% & 6.00\% & 0.00\% & 4.00\% & 10.00\% & 0.00\% & 90.00\% \\
    Time (class 0) & ResNet & 96.0\% & 0.59 $\pm$ 0.26 & 0.00\% & 4.00\% & 0.00\% & 0.00\% & 12.00\% & 0.00\% & 94.00\% \\
    Time (class 1) & ResNet & 100\% & 1.70 $\pm$ 0.18 & 0.00\% & 4.00\% & 0.00\% & 4.00\% & 0.00\% & 0.00\% & 100.00\% \\
    Freq. (class 0) & ResNet & 84.00\%& 0.77 $\pm$ 1.35 & 2.00\% & 2.00\% & 0.00\% & 0.00\% & 20.00\% & 0.00\% & 86.00\% \\
    Freq. (class 1) & ResNet & 100\% & 2.75 $\pm$ 0.99 & 0.00\% & 6.00\% & 0.00\% & 0.00\% & 22.00\% & 0.00\% & 78.00\% \\
    
    \midrule
    Time/Freq. (class 0) & InceptionTime & 100\% & 3.99 $\pm$ 0.81 & 78.00\% & 16.00\% & 0.00\% & 0.00\% & 0.00\% & 0.00\% & 96.00\% \\
    Time/Freq. (class 1) & InceptionTime & 100\% & 2.78 $\pm$ 0.49 & 8.00\% & 4.00\% & 0.00\% & 4.00\% & 26.00\% & 0.00\% & 90.00\% \\
    Time (class 0) & InceptionTime & 100\% & 2.26 $\pm$ 0.43 & 42.00\% & 6.00\% & 14.00\% & 4.00\% & 20.00\% & 0.00\% & 94.00\% \\
    Time (class 1) & InceptionTime & 100\% & 1.89 $\pm$ 0.23 & 44.00\% & 8.00\% & 8.00\% & 0.00\% & 16.00\% & 0.00\% & 92.00\% \\
    Freq. (class 0) & InceptionTime & 96.0\% & 2.17 $\pm$ 0.75 & 6.00\% & 6.00\% & 2.00\% & 2.00\% & 18.00\% & 0.00\% & 82.00\% \\
    Freq. (class 1) & InceptionTime & 100\% & 3.02 $\pm$ 0.19 & 2.00\% & 6.00\% & 2.00\% & 2.00\% & 36.00\% & 0.00\% & 64.00\% \\

    \midrule
    Time/Freq. (class 0) & TCN & 100\% & 2.27 $\pm$ 1.03 & - & 2.00\% & 2.00\% & 2.00\% & 2.00\% & 0.00\% & 100\% \\
    Time/Freq. (class 1) & TCN & 100\% & 1.86 $\pm$ 0.81 & - & 0.00\% & 0.00\% & 0.00\% & 0.00\% & 0.00\% & 100\% \\
    Time (class 0) & TCN & 76.0\% & 0.11 $\pm$ 0.09 & - & 0.00\% & 2.00\% & 0.00\% & 4.00\% & 0.00\% & 100\% \\
    Time (class 1) & TCN & 90.0\% & 0.15 $\pm$ 0.08 & - & 2.00\% & 0.00\% & 0.00\% & 0.00\% & 0.00\% & 100\% \\
    Freq. (class 0) & TCN & 100\% & 2.12 $\pm$ 0.65 & - & 0.00\% & 0.00\% & 0.00\% & 0.00\% & 0.00\% & 100\% \\
    Freq. (class 0) & TCN & 100\% & 1.85 $\pm$ 0.41 & - & 0.00\% & 4.00\% & 2.00\% & 4.00\% & 0.00\% & 100\% \\

    \midrule
    Time/Freq. (class 0) & TST & 90\% & 6.85 $\pm$ 6.09 & 8.00\% & 2.00\% & 8.00\% & 2.00\% & 0.00\% & 0.00\% & 100\% \\
    Time/Freq. (class 1) & TST & 98\% & 7.57 $\pm$ 5.84 & 10.00\% & 8.00\% & 10.00\% & 10.00\% & 2.00\% & 0.00\% & 100\% \\
    Time (class 0) & TST & 38\% & -0.12 $\pm$ 0.81 & 0.00\% & 0.00\% & 0.00\% & 0.00\% & 0.00\% & 0.00\% & 100\% \\
    Time (class 1) & TST & 78\% & 0.35 $\pm$ 0.66 & 6.00\% & 0.00\% & 6.00\% & 0.00\% & 2.00\% & 0.00\% & 100\% \\
    Freq. (class 0) & TST & 98\% & 5.45 $\pm$ 3.57 & 24.00\% & 16.00\% & 24.00\% & 14.00\% & 6.00\% & 0.00\% & 100\% \\
    Freq. (class 0) & TST & 96\% & 5.07 $\pm$ 3.91 & 14.00\% & 12.00\% & 14.00\% & 8.00\% & 2.00\% & 0.00\% & 100\% \\

    \bottomrule
  \end{tabular}
  \caption{UP violation of different XAI methods on the synthetic dataset. For each model, we show the results of different subset of validation separately. For each model, the first two rows show the results for samples that contain both shapelet and frequency features, while the next four rows show the results of samples with only the shapelet or the frequency features, respectively. Activation refers to the models activation (i.e. the last before the SoftMax) of the target class.}
  \label{tbl-synth}
\end{table*}

\textbf{Datasets: } UCR time series repository \cite{dau2019ucr} has been used extensively in time series literature. It contains 128 datasets with various lengths, class numbers, and train/validation sizes. However, due to the nature of the datasets in the UCR repository, we do not expect to see any frequency feature (probably except for FordA or StarLightCurves). For completeness, we include these two datasets.

We use other datasets that potentially contain both time and frequency features, namely MIMIC performance dataset\cite{mimic},
 US hourly climate dataset \cite{ushourly},
 Dhaka Stock Exchange Historical Data \cite{dhaka},
 Valencia air quality (no missing version) from SKForecast repository \cite{SKForecast},
and Monash repository \cite{godahewa2021monash}. MIMIC contains PPG, ECG, and respiration time series. The task is to distinguish adults from the newborns. We subsample each time series multiple times with a fixed length of 1000 time steps. Because each subsample contains around 25 heartbeats, we expect to see a certain frequency profile.

US hourly climate dataset is a public dataset published by the National Centers of Environmental Information. It contains several features recorded at thousands of stations across the United State. Here, we only use the temperature of the Los Angeles airport station (USW00023174). We use the first $90\%$ of the time series as a training set and the rest for validation. The length of each sample is 500 covering almost three weeks of data. This task is a regression of the next day's feature. 

Dhaka stock exchange historical dataset contains stock information of 345 active companies. There is a record for each company per day containing five values: opening price, highest price, lowest price, closing price, and volume. In this paper, we use opening price of the ALLTEX company and the task is forecasting the next day's opening price from the previous 500 observations. We use the first $90\%$ of the time series as a training set and the last $10\%$ for validation.

Valencia air quality contains hourly measures of several air chemical pollutant (i.e. \ce{PM_{2.5}}, \ce{CO}, \ce{NO}, \ce{NO2}, \ce{PM10}, \ce{NOx}, \ce{O3}, air speed, air direction, and \ce{SO2}) in the city of Valencia. Similar to the previous dataset, we forecast \ce{SO2} values based on the previous 500 observations. We use the first $90\%$ of the time series as a training set and the last $10\%$ for validation. 

Monash repository contains $30$ datasets of forecasting task. Here, we choose three with the most potential for having seasonality and trends, namely SF traffic, pedestrian count, and Australia's electricity demand.

Because the ground truth explanations and features of many time series data are not available, we also use a synthetic dataset in this paper to check if the UP violation aligns with ground truth. The synthetic dataset contains four sets of patterns: one distinguishable shapelet feature of length 20 per class, two indistinguishable shapelet randomly added to some samples, one dominant distinguishable frequency feature per class, and an indistinguishable frequency added randomly to some samples. For better clarity, we call indistinguishable patterns \textit{non-features} in the rest of the paper. During training, we force half of the training samples to only have distinguishable shapelet features and the other half to only have distinguishable frequency features. This ensures that the model learns both frequency and time features if it achieves high training accuracy.
The non-features are randomly added to any of the samples. The test samples are generated into three groups: those with only the distinguishable frequency features, those with only shapelet features, and those with both. A desirable XAI method should not violate the UP in the first two groups, while UP violation in the third group indicates the ability of an XAI method to localize different features in different domains.

\begin{table*}[ht]
\scriptsize
  \centering
  \begin{tabular}{llllllllll}
    \toprule
    Dataset & Model & Accuracy/MSE & DeepLIFT & GradSHAP & I $\times$ G & IG & Occlusion & Saliency & LIME\\
    \midrule
    MIMICPerform & FCN & 89.50\% & 0.00\% & 0.00\% & 0.00\% & 0.00\% & 0.00\% & 0.00\% & 75.00\% \\
    MIMICPerformECG & FCN & 77.82\% & 0.00\% & 0.00\% & 0.00\% & 0.00\% & 6.00\% & 0.00\% & 100\% \\
    MIMICPerformRESP & FCN & 91.53\% & 0.00\% & 0.00\% & 0.00\% & 0.00\% & 0.00\% & 0.00\% & 100\% \\
    FordA & FCN & 79.70\% & 0.00\% & 0.00\% & 0.00\% & 0.00\% & 0.00\% & 0.00\% & 95.00\% \\
    StarLightCurves & FCN & 92.19\% & 11.00\% & 9.00\% & 11.00\% & 10.00\% & 10.00\% & 0.00\% & 99.00\% \\

    \midrule
    USHourlyClimate & FCN & 4.88e-05 & 0.00\% & 1.00\% & 0.00\% & 1.00\% & 0.00\% & 0.00\% & 100\% \\
    DhakaStockExchange & FCN & 6.62e-05 & 0.00\% & 0.00\% & 0.00\% & 0.00\% & 0.00\% & 0.00\% & 99.00\% \\
    SKForecastAirQuality & FCN & 7.22e-03 & 0.00\% & 0.00\% & 0.00\% & 0.00\% & 0.00\% & 0.00\% & 98.00\% \\
    MonashSFTraffic & FCN & 3.50e-06 & 79.00\% & 77.00\% & 75.00\% & 79.00\% & 76.00\% & 0.00\% & 64.00\% \\
    MonashPedestrianCounts & FCN & 9.35e-05 & 38.00\% & 13.00\% & 36.00\% & 11.00\% & 42.00\% & 0.00\% & 88.00\% \\
    MonashAusElectricityDemand & FCN & 6.95e-05 & 53.00\% & 57.00\% & 53.00\% & 58.00\% & 55.00\% & 0.00\% & 100\% \\

    \midrule
    MIMICPerform & ResNet & 95.57\% & 4.00\% & 1.00\% & 17.00\% & 0.00\% & 0.00\% & 0.00\% & 100\% \\
    MIMICPerformECG & ResNet & 98.25\% & 0.00\% & 0.00\% & 1.00\% & 0.00\% & 0.00\% & 0.00\% & 100\% \\
    MIMICPerformRESP & ResNet & 93.44\% & 0.00\% & 0.00\% & 1.00\% & 1.00\% & 0.00\% & 0.00\% & 100\% \\
    FordA & ResNet & 94.47\% & 0.00\% & 0.00\% & 0.00\% & 0.00\% & 0.00\% & 0.00\% & 99.00\% \\
    StarLightCurves & ResNet & 97.63\% & 7.00\% & 7.00\% & 0.00\% & 7.00\% & 0.00\% & 0.00\% & 99.00\% \\

    \midrule
    USHourlyClimate & ResNet & 3.98e-03 & 46.00\% & 40.00\% & 45.00\% & 45.00\% & 0.00\% & 0.00\% & 89.00\% \\
    DhakaStockExchange & ResNet & 1.69e-04 & 100\% & 78.00\% & 99.00\% & 100\% & 0.00\% & 0.00\% & 100\% \\
    SKForecastAirQuality & ResNet & 1.10e-02 & 23.00\% & 3.00\% & 2.00\% & 5.00\% & 23.00\% & 0.00\% & 97.00\% \\
    MonashSFTraffic & ResNet & 1.51e-04 & 88.00\% & 50.00\% & 51.00\% & 58.00\% & 40.00\% & 0.00\% & 91.00\% \\
    MonashPedestrianCounts & ResNet & 3.99e-03 & 58.00\% & 25.00\% & 13.00\% & 24.00\% & 24.00\% & 0.00\% & 67.00\% \\
    MonashAusElectricityDemand & ResNet & 2.82e-03 & 89.00\% & 89.00\% & 77.00\% & 95.00\% & 70.00\% & 0.00\% & 100\% \\

    \midrule
    MIMICPerform & InceptionTime & 96.19\% & 2.00\% & 0.00\% & 23.00\% & 0.00\% & 0.00\% & 0.00\% & 100\% \\
    MIMICPerformECG & InceptionTime & 95.33\% & 10.00\% & 4.00\% & 0.00\% & 1.00\% & 0.00\% & 0.00\% & 97.00\% \\
    MIMICPerformRESP & InceptionTime & 93.42\% & 0.00\% & 1.00\% & 0.00\% & 1.00\% & 0.00\% & 0.00\% & 100\% \\
    FordA & InceptionTime & 93.41\% & 100\% & 0.00\% & 0.00\% & 0.00\% & 0.00\% & 0.00\% & 97.00\% \\
    StarLightCurves & InceptionTime & 96.78\% & 0.00\% & 0.00\% & 0.00\% & 1.00\% & 0.00\% & 0.00\% & 100\% \\

    \midrule
    USHourlyClimate & InceptionTime & 4.89e-04 & 0.00\% & 13.00\% & 0.00\% & 21.00\% & 0.00\% & 0.00\% & 100\% \\
    DhakaStockExchange & InceptionTime & 2.83e-05 & 0.00\% & 50.00\% & 61.00\% & 100\% & 0.00\% & 0.00\% & 100\% \\
    SKForecastAirQuality & InceptionTime & 8.64e-03 & 5.00\% & 1.00\% & 1.00\% & 2.00\% & 1.00\% & 0.00\% & 98.00\% \\
    MonashSFTraffic & InceptionTime & 4.09e-06 & 0.00\% & 49.00\% & 6.00\% & 55.00\% & 26.00\% & 0.00\% & 79.00\% \\
    MonashPedestrianCounts & InceptionTime & 1.06e-04 & 3.00\% & 14.00\% & 19.00\% & 16.00\% & 32.00\% & 0.00\% & 76.00\% \\
    MonashAusElectricityDemand & InceptionTime & 7.29e-05 & 51.00\% & 75.00\% & 72.00\% & 84.00\% & 22.00\% & 0.00\% & 100\% \\

    \midrule
    MIMICPerform & TCN & 93.93\% & - & 14.00\% & 6.00\% & 3.00\% & 1.00\% & 0.00\% & 99.00\% \\
    MIMICPerformECG & TCN & 94.48\% & - & 1.00\% & 0.00\% & 0.00\% & 0.00\% & 0.00\% & 93.00\% \\
    MIMICPerformRESP & TCN & 92.26\% & - & 0.00\% & 0.00\% & 0.00\% & 0.00\% & 0.00\% & 99.00\% \\
    FordA & TCN & 91.67\% & - & 0.00\% & 0.00\% & 0.00\% & 0.00\% & 0.00\% & 98.00\% \\
    StarLightCurves & TCN & 85.53\% & - & 2.00\% & 0.00\% & 2.00\% & 0.00\% & 0.00\% & 100\% \\

    \midrule
    USHourlyClimate & TCN & 3.04e-03 & - & 28.00\% & 25.00\% & 32.00\% & 20.00\% & 0.00\% & 100\% \\
    DhakaStockExchange & TCN & 5.50e-04 & - & 0.00\% & 0.00\% & 0.00\% & 0.00\% & 0.00\% & 84.00\% \\
    SKForecastAirQuality & TCN & 1.08e-02 & - & 5.00\% & 0.00\% & 5.00\% & 0.00\% & 0.00\% & 100\% \\
    MonashSFTraffic & TCN & 1.05e-05 & - & 0.00\% & 0.00\% & 0.00\% & 0.00\% & 0.00\% & 100\% \\
    MonashPedestrianCounts & TCN & 2.02e-04 & - & 8.00\% & 3.00\% & 9.00\% & 4.00\% & 0.00\% & 74.00\% \\
    MonashAusElectricityDemand & TCN & 4.36e-03 & - & 38.00\% & 45.00\% & 39.00\% & 35.00\% & 0.00\% & 100\% \\

    \midrule
    MIMICPerform & TST & 86.80\% & 0.00\% & 0.00\% & 0.00\% & 0.00\% & 0.00\% & 0.00\% & 99.00\% \\
    MIMICPerformECG & TST & 70.79\% & 2.00\% & 3.00\% & 2.00\% & 3.00\% & 1.00\% & 0.00\% & 100\% \\
    MIMICPerformRESP & TST & 81.06\% & 0.00\% & 0.00\% & 0.00\% & 0.00\% & 0.00\% & 0.00\% & 89.00\% \\
    StarLightCurves & TST & 89.87\% & 0.00\% & 3.00\% & 0.00\% & 4.00\% & 0.00\% & 0.00\% & 100\% \\

    \midrule
    USHourlyClimate & TST & 6.81e-04 & 0.00\% & 0.00\% & 0.00\% & 0.00\% & 0.00\% & 0.00\% & 100\% \\
    DhakaStockExchange & TST & 3.44e-03 & 0.00\% & 0.00\% & 0.00\% & 0.00\% & 0.00\% & 0.00\% & 100\% \\
    SKForecastAirQuality & TST & 1.50e-02 & 3.00\% & 7.00\% & 3.00\% & 7.00\% & 4.00\% & 0.00\% & 100\% \\
    MonashSFTraffic & TST & 8.85e-06 & 39.00\% & 2.00\% & 39.00\% & 1.00\% & 3.00\% & 0.00\% & 100\% \\
    MonashPedestrianCounts & TST & 8.49e-05 & 37.00\% & 42.00\% & 37.00\% & 41.00\% & 47.00\% & 0.00\% & 91.00\% \\
    MonashAusElectricityDemand & TST & 8.47e-04 & 57.00\% & 52.00\% & 57.00\% & 48.00\% & 58.00\% & 0.00\% & 99.00\% \\

    \bottomrule
  \end{tabular}
  \caption{UP violation of different XAI methods on real datasets.}
  \label{tbl-real}
\end{table*}

\begin{figure*}[ht]
    \centering
    \subfloat[MIMICPerform (Time)]{%
        \includegraphics[width=0.23\linewidth, height=0.12\textwidth]{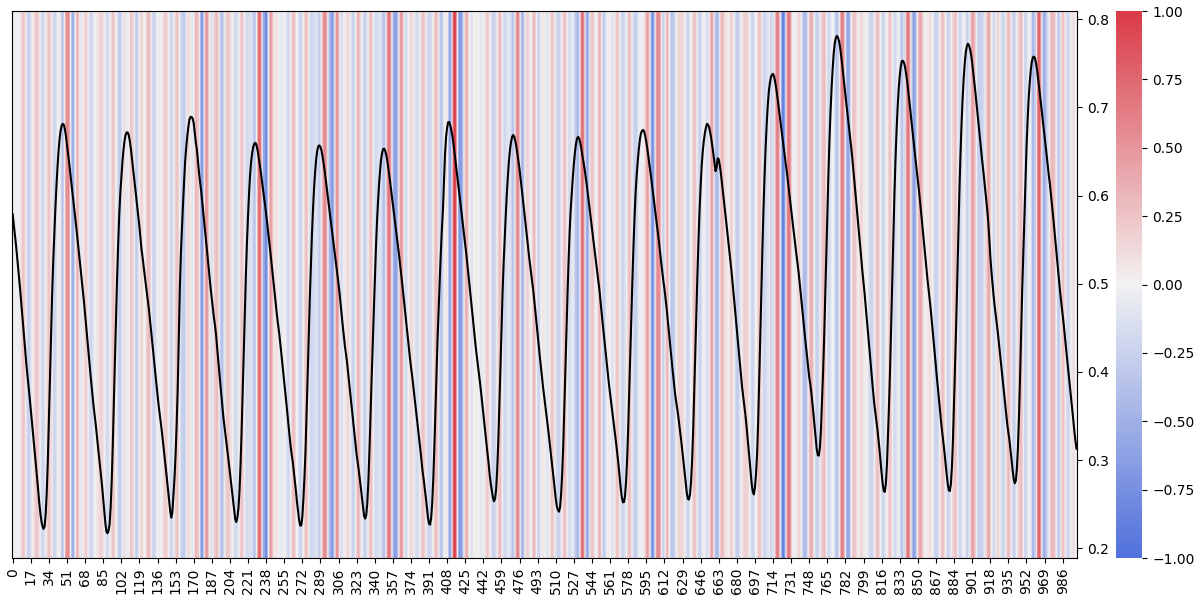}
        }
    \subfloat[MIMICPerform (Freq.)]{
        \includegraphics[width=0.23\linewidth, height=0.12\textwidth]{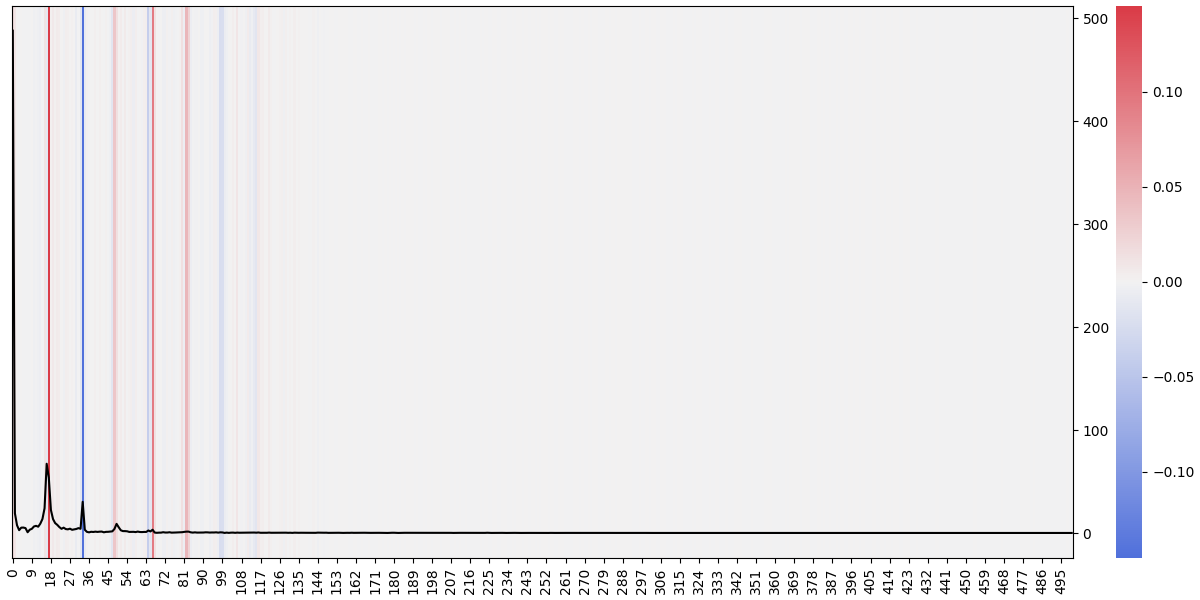}
        }
    \subfloat[MIMICPerform (Time)]{%
        \includegraphics[width=0.23\linewidth, height=0.12\textwidth]{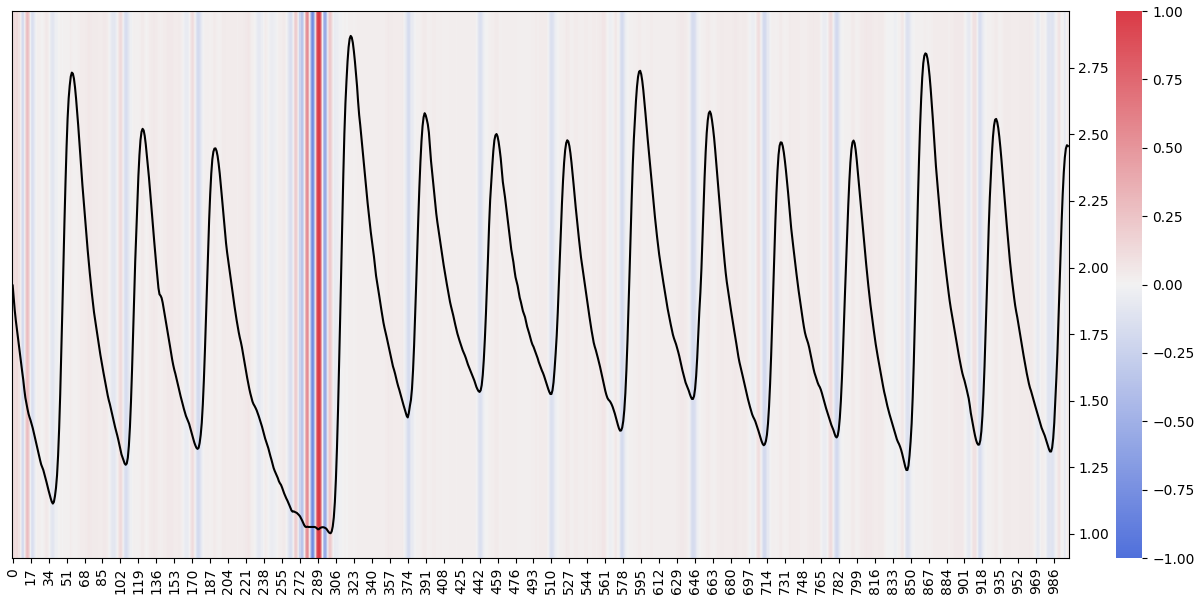}
        }
    \subfloat[MIMICPerform (Freq.)]{
        \includegraphics[width=0.23\linewidth, height=0.12\textwidth]{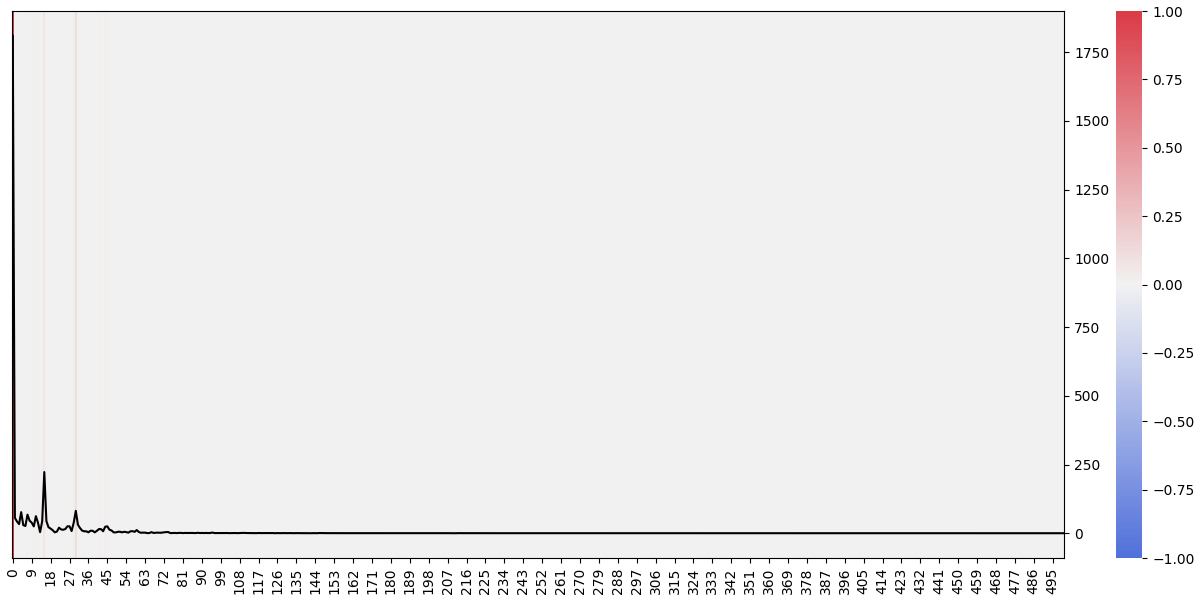}
        }
    \\
    \subfloat[PedestrianCount (Time)]{%
        \includegraphics[width=0.23\linewidth, height=0.12\textwidth]{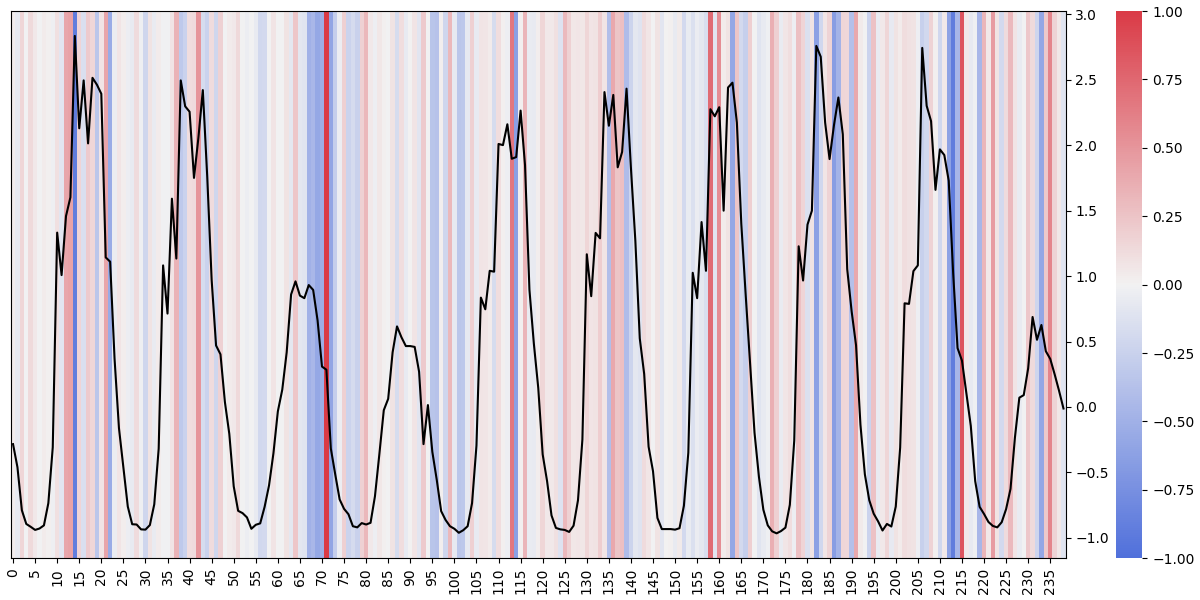}
        }
    \subfloat[PedestrianCount (Freq.)]{
        \includegraphics[width=0.23\linewidth, height=0.12\textwidth]{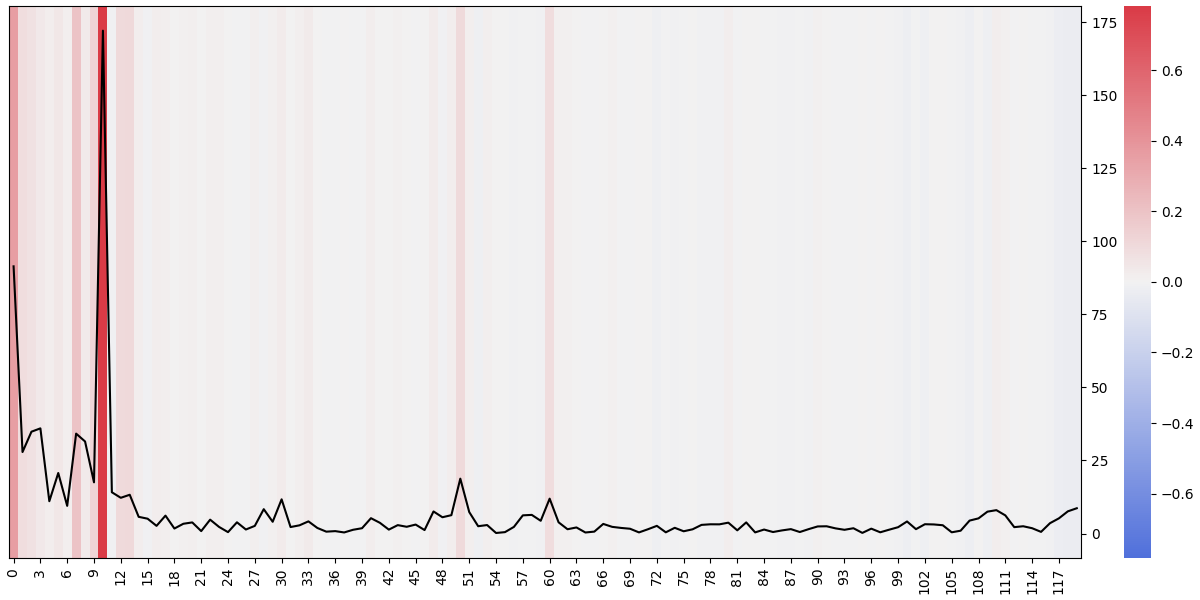}
        }
    \subfloat[PedestrianCount (Time)]{%
        \includegraphics[width=0.23\linewidth, height=0.12\textwidth]{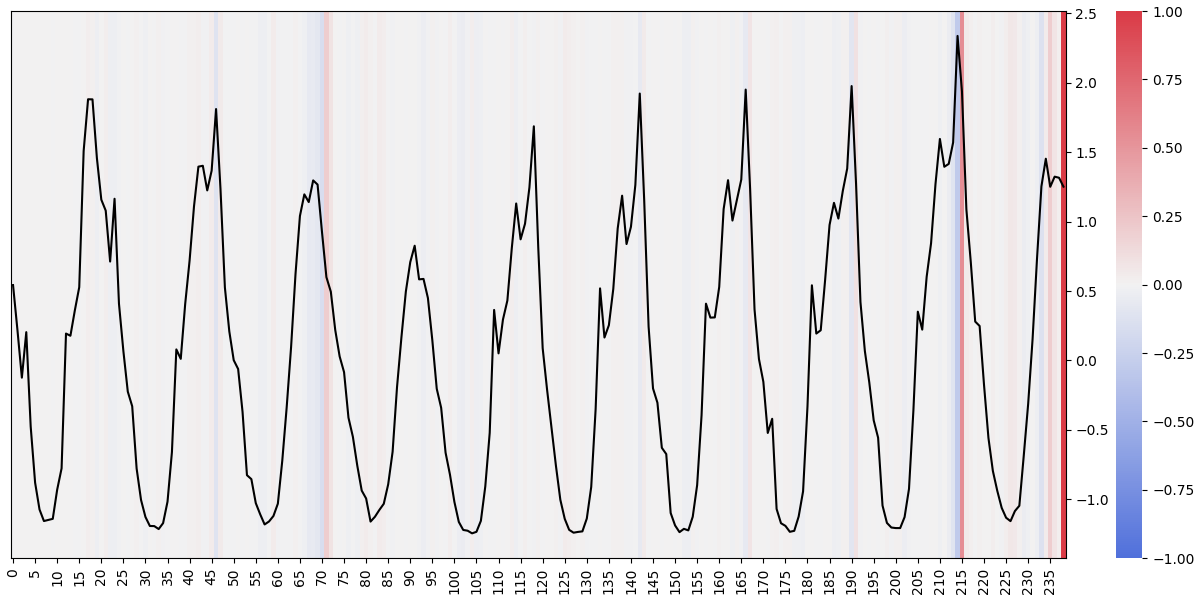}
        }
    \subfloat[PedestrianCount (freq.)]{
        \includegraphics[width=0.23\linewidth, height=0.12\textwidth]{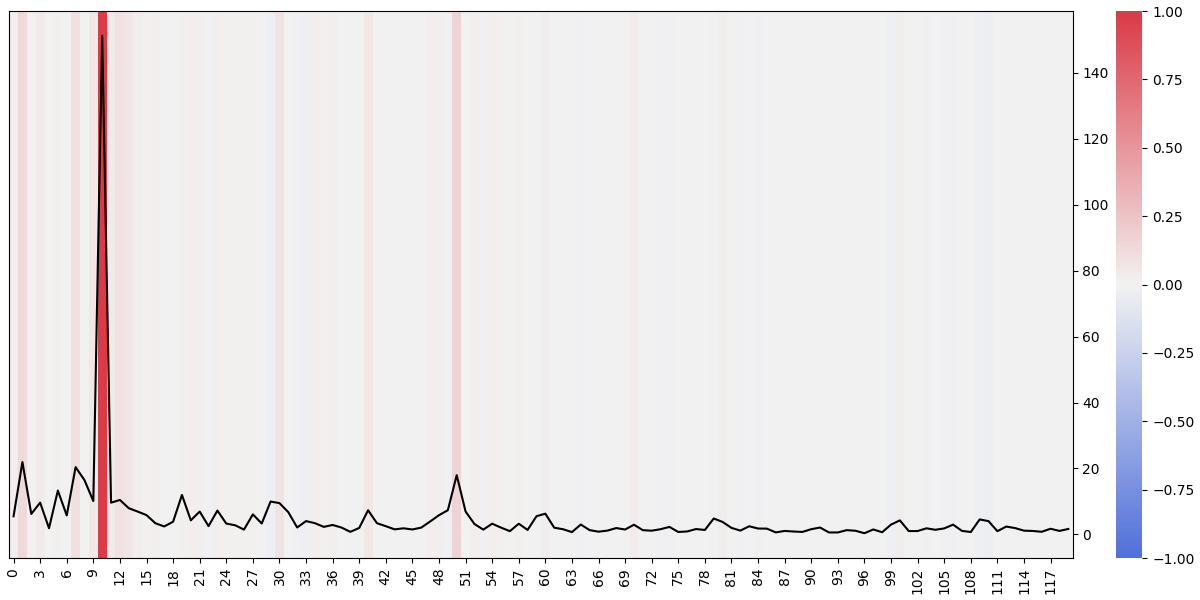}
        }
    \\
    \subfloat[USHourlyClimate (Time)]{%
        \includegraphics[width=0.23\linewidth, height=0.12\textwidth]{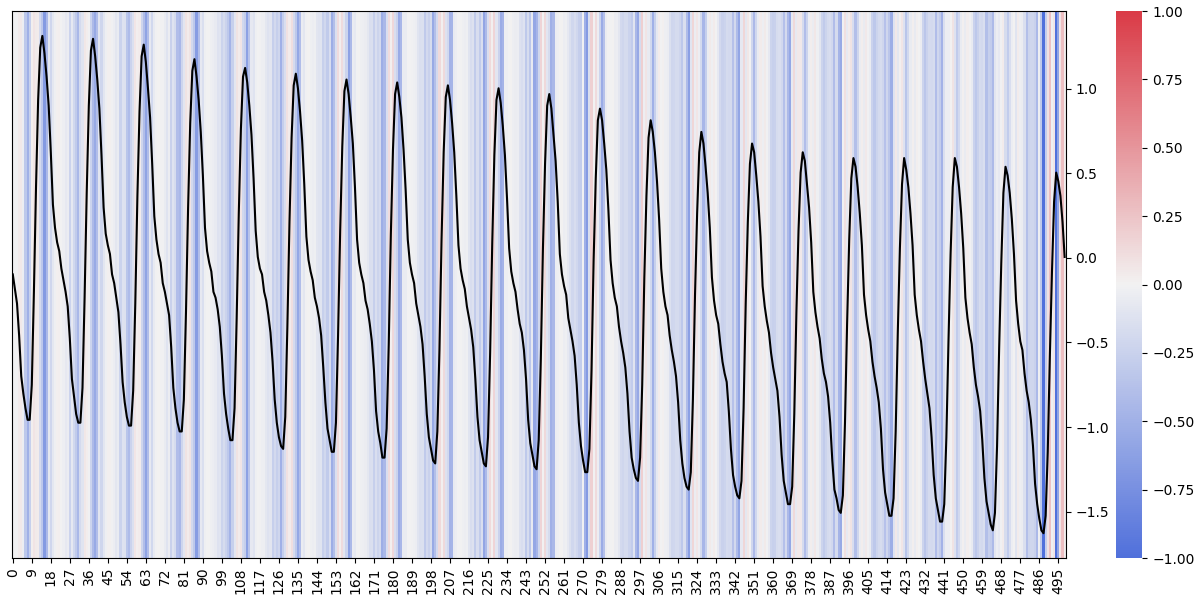}
        }
    \subfloat[USHourlyClimate (Freq.)]{
        \includegraphics[width=0.23\linewidth, height=0.12\textwidth]{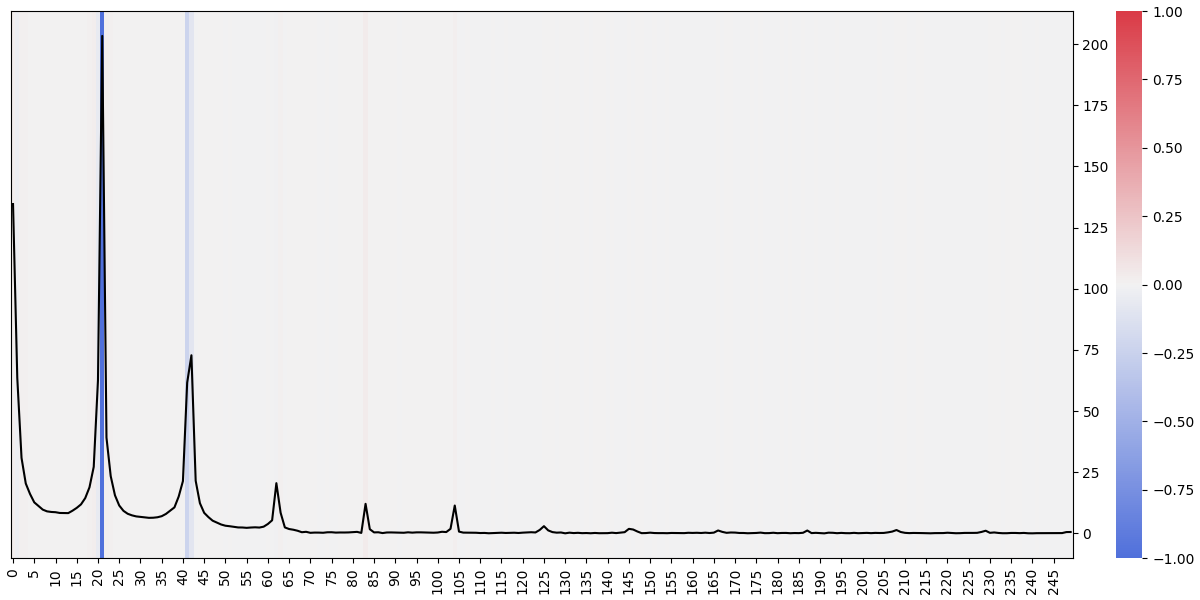}
        }
    \subfloat[USHourlyClimate (Time)]{%
        \includegraphics[width=0.23\linewidth, height=0.12\textwidth]{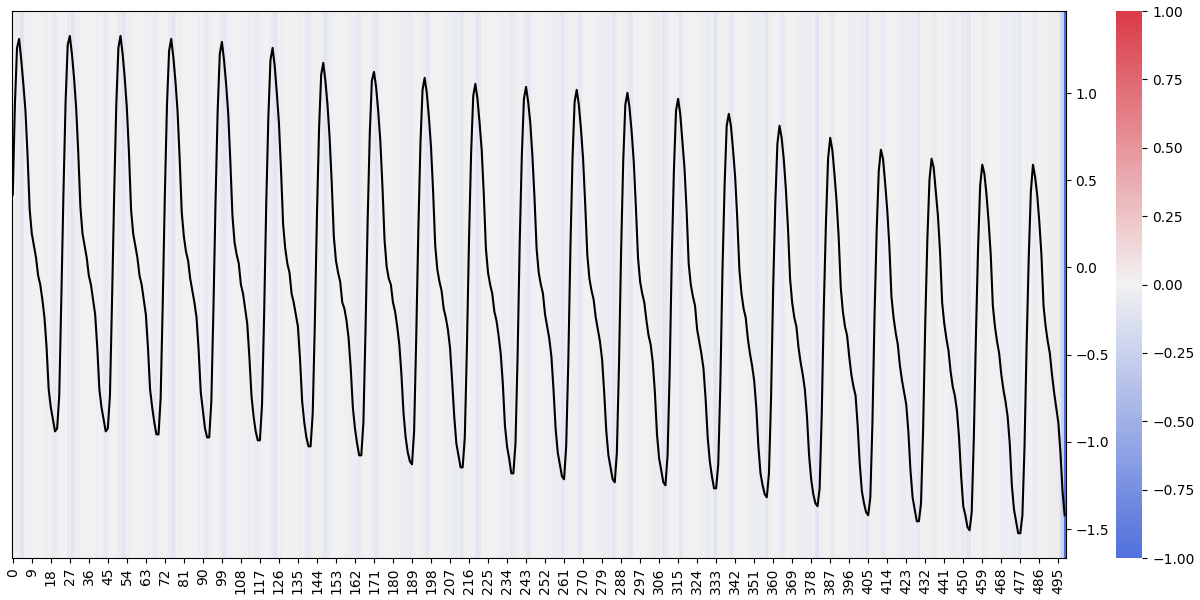}
        }
    \subfloat[USHourlyClimate (Freq.)]{
        \includegraphics[width=0.23\linewidth, height=0.12\textwidth]{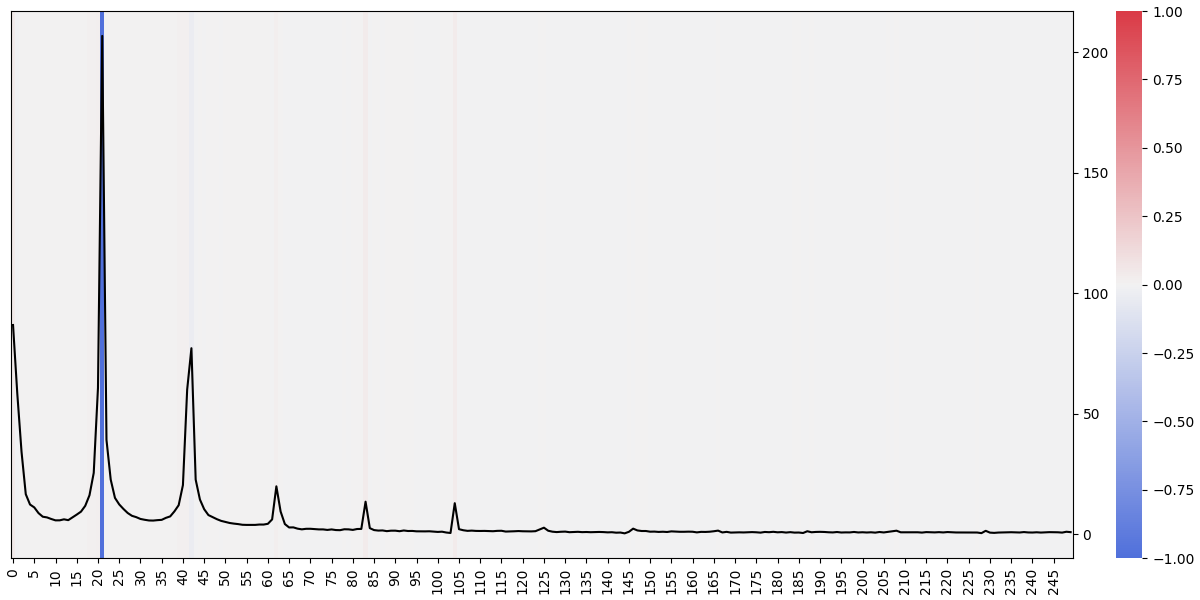}
        }
    \\
    \subfloat[StarLightCurve (Time)]{%
        \includegraphics[width=0.23\linewidth, height=0.12\textwidth]{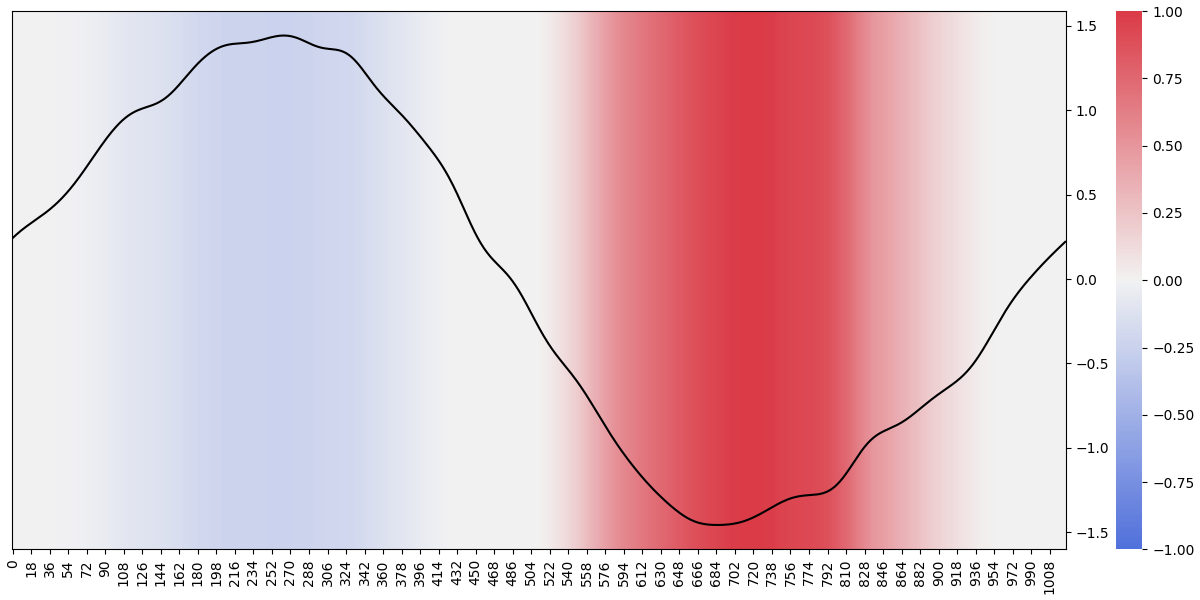}
        }
    \subfloat[StarLightCurve (Freq.)]{
        \includegraphics[width=0.23\linewidth, height=0.12\textwidth]{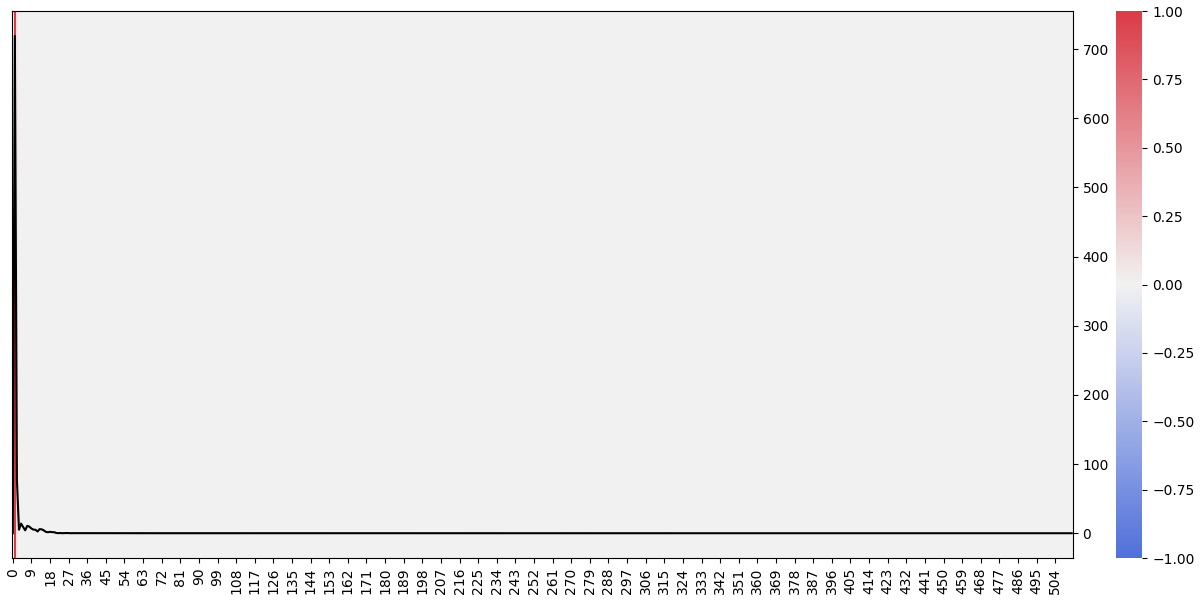}
        }
    \subfloat[StarLightCurve (Time)]{%
        \includegraphics[width=0.23\linewidth, height=0.12\textwidth]{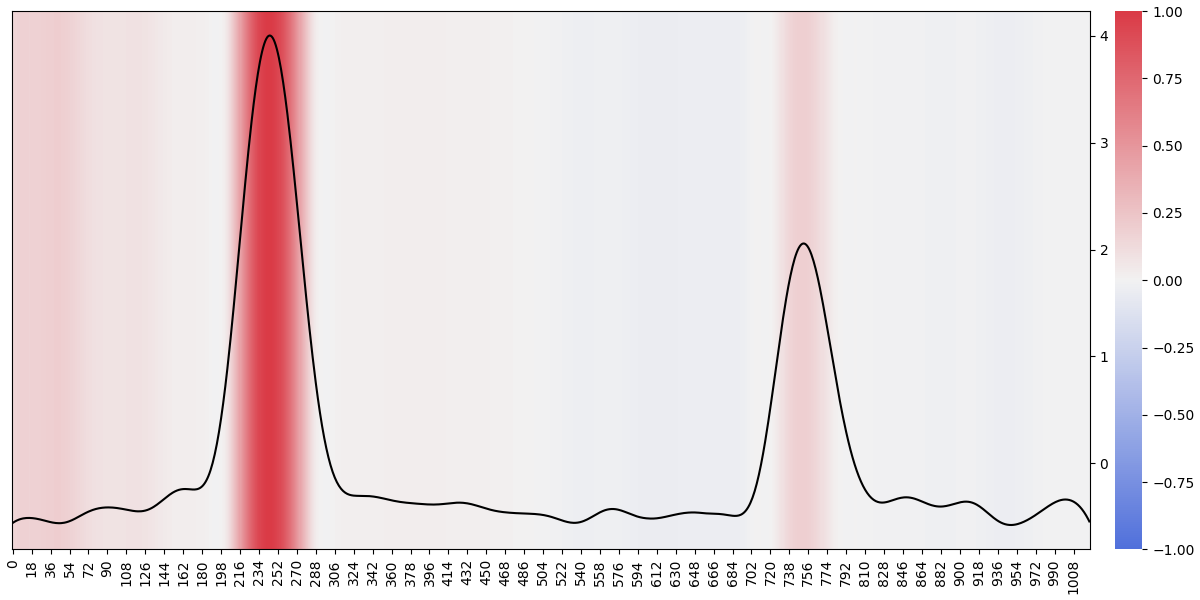}
        }
    \subfloat[StarLightCurve (Freq.)]{
        \includegraphics[width=0.23\linewidth, height=0.12\textwidth]{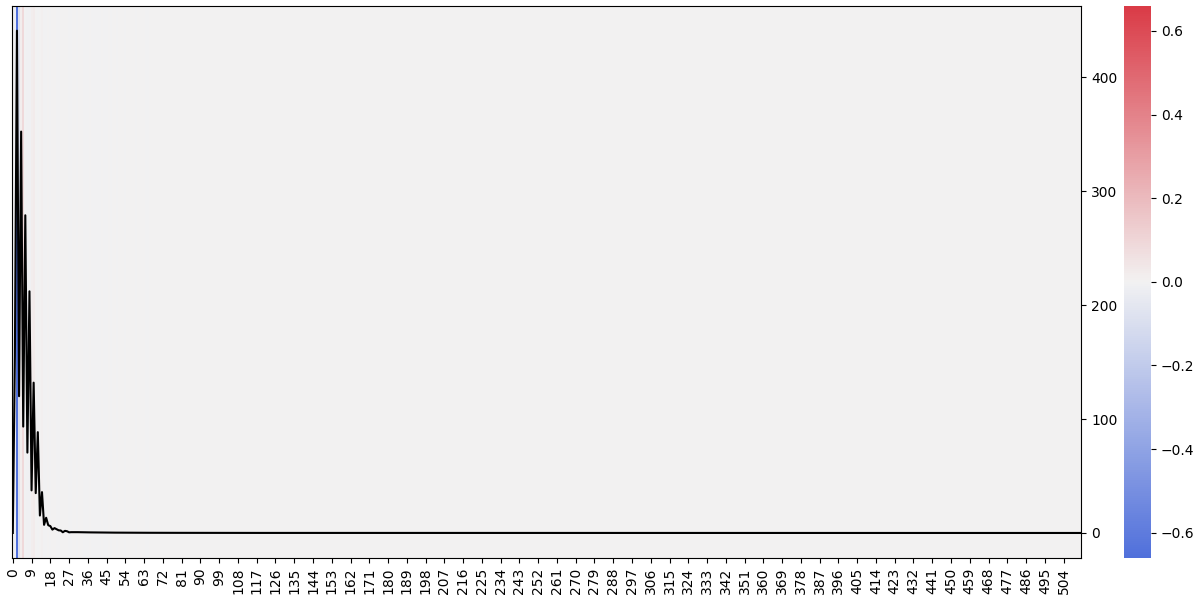}
        }
    \caption{UP violation on real datasets: The first two columns and the last two columns are time frequency pair of a same sample. The fist two columns are examples where UP is not violated and, consequently, time and frequency attributions are not both localized. The last two columns are examples where the UP is violated and, hence, the attributions cannot be associated to the same feature. The first row shows the attributions of DeepLIFT on a ResNet model. The second row illustrates the attributions of occlusion method on a FCN model. The third row demonstrates the attributions of GradientSHAP on a TST model. The fourth row shows the attributions of Integrated Gradient on an TCN model. The last row illustrates the attributions of Integrated Gradient on a ResNet model. 
    }
    \label{fig:real-examples}
\end{figure*}

\subsection{UP Violation Results}
The results of the uncertainty principle violations are presented in Tables \ref{tbl-synth} and \ref{tbl-real}. 
One interesting observation is that the results of different models and XAI methods on the same dataset/model are not consistent. First, LIME method violate UP in most cases. This has been observed in \cite{rezaei2024explanation} and it is due to the sparsity regularization enforced by LIME. Hence, the attributions generated by LIME are always sparse regardless of the explanation domain and the dataset. This issue can be mitigated by aggregating multiple explanations, as we will discuss in Section \ref{sec:lime}. We ignore the results of LIME here. Second, the only XAI method that never violates UP is Saliency. We speculate that it is due to the fact that Saliency does not have any implicit or explicit baseline. Focusing on the simple case of the occlusion method, distinguishable frequency feature in the frequency domain, if zeroed out, has a huge impact on the time domain and, consequently, on the model's prediction. This is not true for other non-feature frequencies with small amplitude. On the other hand, a shapelet feature is distributed among several frequency components of the frequency domain, and they all have very small amplitude. Hence, zeroing out one does not significantly affect the input. So, the methods that have a baseline favor the features which have a larger difference with the baseline which, in the case of frequency domain, may cause the attribution to highlight the high amplitude frequency feature. However, the Saliency method does not have a baseline or zeroing out process. It computes the gradient with respect to the input. Hence, the magnitude of the original input value and its difference to a baseline does not affect the attribution. As a result, it shows both the frequency feature and the distributed components of the shapelet. Therefore, Saliency does not localize attributions in the frequency and time domain at the same time. 

We start with the synthetic dataset in Table \ref{tbl-synth}. Based on the ground truth, we expect to see potential UP violation on samples with both time and frequency features and not on samples with only time or frequency features, if deep models are perfectly aligned with ground truth and XIA methods are faithful. However, there are some exceptions to this rule with InceptionTime model on samples with time features being the most outlier. We will explain the root cause of this problem in Section \ref{sec:amp-fre-response}. An example of synthetic samples with DeepLift attribution is provided in Figure \ref{fig:synthetic}, where the attributions in the time and frequency domains are both localized when both features exist. For ResNet, InceptionTime and TST, we see UP violations captured by various methods, including DeepLift, GradientSHAP, and Integrated Gradient. Interestingly, TCN and TST struggle with time domain features and could not achieve high accuracy. In TST, we observe that the UP voilation on samples with only frequency features are caused by the fact that the TST model associates one of the time-domain non-features with one class and that become highlighted by some XIA methods. Another reason we see fewer UP violations in FCN, TCN, and TST is evident by looking at their activation values.
For these models, the activations are significantly larger for samples with frequency features compared to samples with shapelet features. As a result, for samples that contain both features, the contribution of frequency to the activation is much larger, and consequently, XAI methods show the frequency feature even in the time domain. That is why we see much fewer UP violation on these models. Such activation values can also explain why the UP violation occurs more often in ResNet and InceptionTime. For example, in ResNet, the average activation values of class 0 samples that only contain shapelet is $0.59\pm0.26$ which is close to the samples with only frequency feature ($0.77\pm1.35$). Because one feature is not dominant in terms of its effect on model's output, some XAI methods can highlight the sparser feature in its corresponding domain, i.e. highlighting shapelet in time domain and dominant frequency in frequency domain. Hence, the UP violation occurs. 

We emphasize that \textit{the dataset (ground truth) features of our synthetic dataset are not necessarily aligned with what different models learned, nor with what an XAI method can bring up}. This is well known in the literature \cite{ismail2020benchmarking, loffler2022don}. How exactly each individual XAI method and model architecture affect attribution generation in different domains is beyond the scope of this paper. However, when UP is violated, regardless of what the ground truth is, it is a sufficient condition that time and frequency domain attributions do not highlight the same feature. For this reason, UP violation in samples with only time (frequency) feature is not wrong, i.e. not a false positive. In other words, UP violation means attributions that do not highlight the same feature, whether they align with ground truth or not. This seeming contradiction can stem form from DL models or XAI methods.

Figure \ref{fig:real-examples} demonstrates the attributions of various XAI methods on different datasets and models. The first two columns show the time/frequency pairs of the same samples in which the UP is not violated. The last two columns illustrate the time/frequency pairs of the same samples in which the UP is violated. It is clear that the attributions on the left-hand side are not localized in the time and frequency domains, simultaneously, while the last two pairs' attributions are localized in both domains. Despite the lack of ground truth explanation for real datasets, it is reasonable to expect that forecast tasks rely mostly on the last time step and some seasonality feature if exist. In MonashPedestrianCount and USHourlyClimate, which are both forecast tasks, time domain attribution shows the importance of the last time step (Figure \ref{fig:real-examples} (g) and (k)), while the frequency domain shows the importance of a certain frequency (Figure \ref{fig:real-examples} (h) and (l)). This example clearly manifests the limitation of widely-used XAI methods on time domain: Although both features are used by the model, attribution in only one domain cannot demonstrate this fact to the end user. Another example is the MIMICPerform dataset (Figure \ref{fig:real-examples} (a) to (d)). Interestingly, we find that, unlike the majority of the time series data we use here, the MIMICPerformance is not per-sample z-normalized. More interestingly, the mean values of training samples for the two classes are considerably different. What we see in Figure \ref{fig:real-examples} (c) and (d) where UP violation occurs is that the time domain shows the shapelet anomaly while the frequency domain shows the mean value that the model exploits.

\begin{figure}[]
    \centering
    \subfloat[Time]{%
        \includegraphics[width=0.70\linewidth, height=0.12\textwidth]{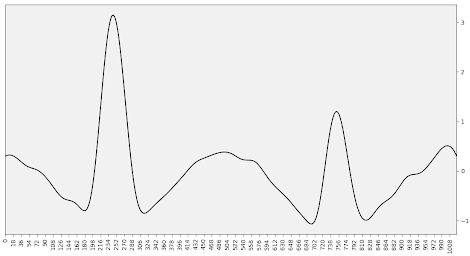}
        }
    \caption{The result of removing the negatively highlighted frequency in Figure \ref{fig:real-examples} (p) when converted back to time domain which shows the appearance of crest similar to the other class.}
    \label{fig:starcurve-modified}
\end{figure}

Among all the classification tasks, the StarLightCurve is among the few classification tasks with solid physics ground truth supporting the observation \cite{rebbapragada2009finding}. This dataset contains the magnitude of the brightness of different celestial objects. In astronomy observations, the magnitude is inversely proportional to the brightness of the observation, thus, the greater the y-axis the dimmer the object is \cite{rebbapragada2009finding}. Figures \ref{fig:real-examples} (m) and (n) show a light curve of a Cepheid variable star (class 0), while Figures \ref{fig:real-examples} (o) and (p) show the light curve of eclipsing binary stars. The pulsation of Cepheid variables are well-known through the kappa mechanism \cite{engle2015secret}. Kappa mechanism is responsible for compression and expansion of a variable star which results in a single sine-wave time series in each expansion/compression period. The smoothness of the time series in StarLightCurve is due to the averaging mechanism over multiple observation periods of the same object, called \textit{folding} \cite{rebbapragada2009finding}. The attribution in Figure \ref{fig:real-examples} (m) almost covers the entire time series and the corresponding attribution in frequency domain in Figure \ref{fig:real-examples} (n) clearly shows that the model picks the lowest frequency component (after the zero component corresponding to the average value). Eclipsing binaries refer to two starts that orbit each others. When the orbital plane of the two stars is nearly aligned with the observer's line of sight, the stars undergo an eclipse \cite{kallrath2009eclipsing}. Hence, the light of one start is obscured by the one in front of the observer, and the total light received by the observer decreases. This happens twice in each complete period. We can see the two eclipses in Figure \ref{fig:real-examples} (o) at the two peaks highlighted by Integrated Gradient. Frequency attribution in Figure \ref{fig:real-examples} (p) contains one highly negative frequency responsible for interfering with other frequencies to zero out the rest of the time domain non-peak regions. We can see the result of removing this frequency in Figure \ref{fig:starcurve-modified} on the time domain. This causes crest-like features to appear which is similar to features of the Cepheid variable class. This clearly shows that the attributions highlighted in Figure \ref{fig:real-examples} (o) and (p) are completely different. In summary, all these examples demonstrate the importance of multi-domain explanations and the usefulness of uncertainty principle as a sufficient condition for multi-domain explanations.

\subsection{LIME Issue}
\label{sec:lime}

\begin{figure}[]
    \centering
    \subfloat[Time]{%
        \includegraphics[width=0.45\linewidth, height=0.12\textwidth]{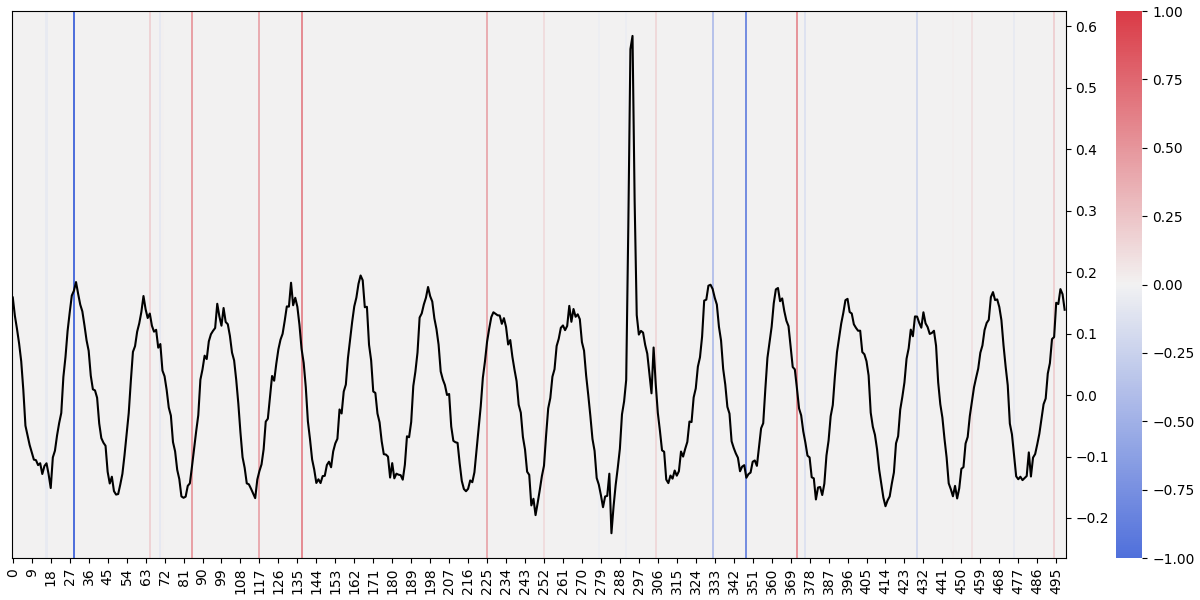}
        }
    \subfloat[Frequency]{
        \includegraphics[width=0.45\linewidth, height=0.12\textwidth]{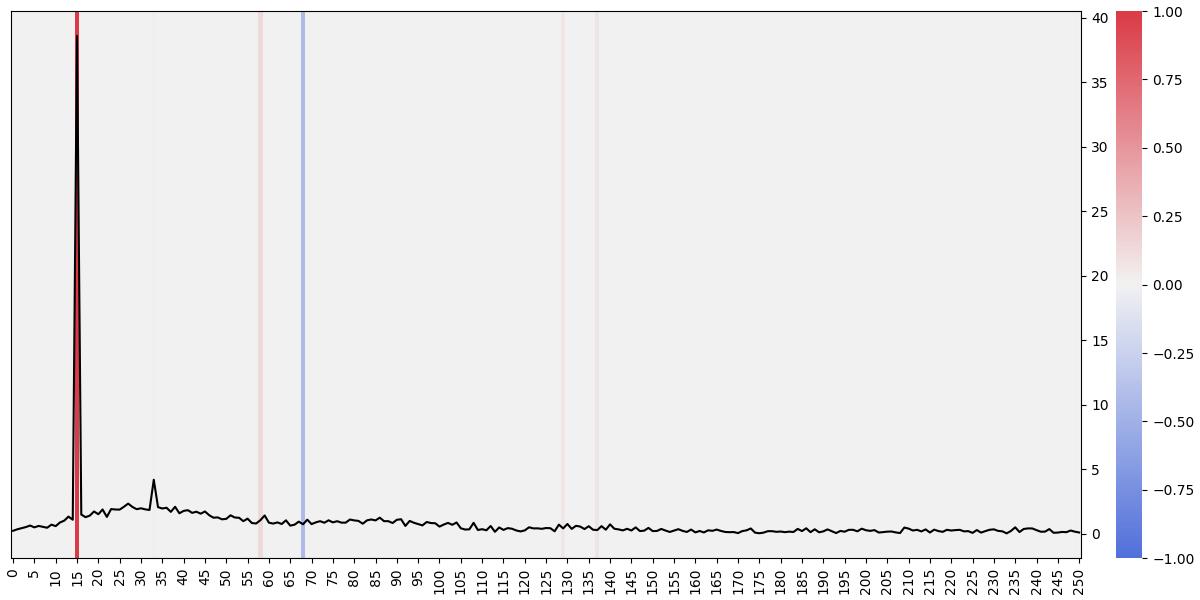}
        }
    \\
    \subfloat[Time]{%
        \includegraphics[width=0.45\linewidth, height=0.12\textwidth]{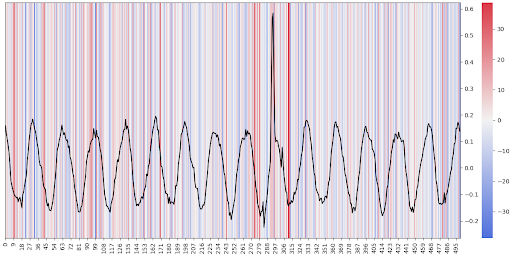}
        }
    \subfloat[Frequency]{
        \includegraphics[width=0.45\linewidth, height=0.12\textwidth]{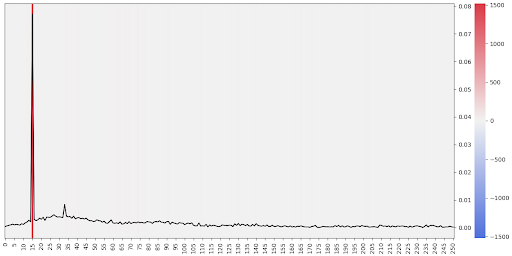}
        }
    \caption{LIME attribution on synthetic dataset containing both shapelet features and dominant frequency components. The first row shows the original attribution from LIME while the second row illustrates the aggregated attributions of 100 LIME's output generated from running LIME with different random seeds. While the first row attribution pair violates uncertainty principle, the second row pair does not.}
    \label{fig:lime}
\end{figure}

The LIME method has a specific regularization in which it enforces sparsity. LIME trains a simpler interpretable linear surrogate model that only takes $k$ features (i.e. input time steps). This explicit sparsity regularization is the reason for the attribution in both time and frequency domain to be sparse which, consequently, causes uncertainty principle violation. For consistency, we use an off-the-shelf XAI implementation of Captum \cite{kokhlikyan2020captum} in which there is no easy way to allow $k$ to be as large as the entire input. In fact, this may even defy the original purpose of the method. However, in time series, there are cases where the represented feature is not sparse at all, e.g. a dominant frequency represented in time domain. In this case, we recommend using LIME multiple times on the same data point and aggregate the results. For cases where the feature is localized, no change will occur after aggregation. For cases where the feature is not localized, however, each run of the LIME picks a subset of the entire feature. Hence, aggregating them will reconstruct the non-localized attribution. Figure \ref{fig:lime} demonstrates an example of the aggregation method on a sample with the frequency feature of our synthetic dataset. Original LIME attribution, shown in \ref{fig:lime} (a) and (b), is very sparse and, more importantly, is very misleading in the time domain. Moreover, it inevitably highlights some unrelated frequency domain artifacts. Aggregating 100 iterations of LIME, shown in \ref{fig:lime} (c) and (d), removes the frequency domain's artifacts and reverts the sparsity regulation in the time domain so that it shows the entire sine wave.

\subsection{Impact of DL Models on Frequency Representation}
\label{sec:amp-fre-response}
\begin{figure}[]
    \centering
    \subfloat[Time]{%
        \includegraphics[width=0.45\linewidth, height=0.12\textwidth]{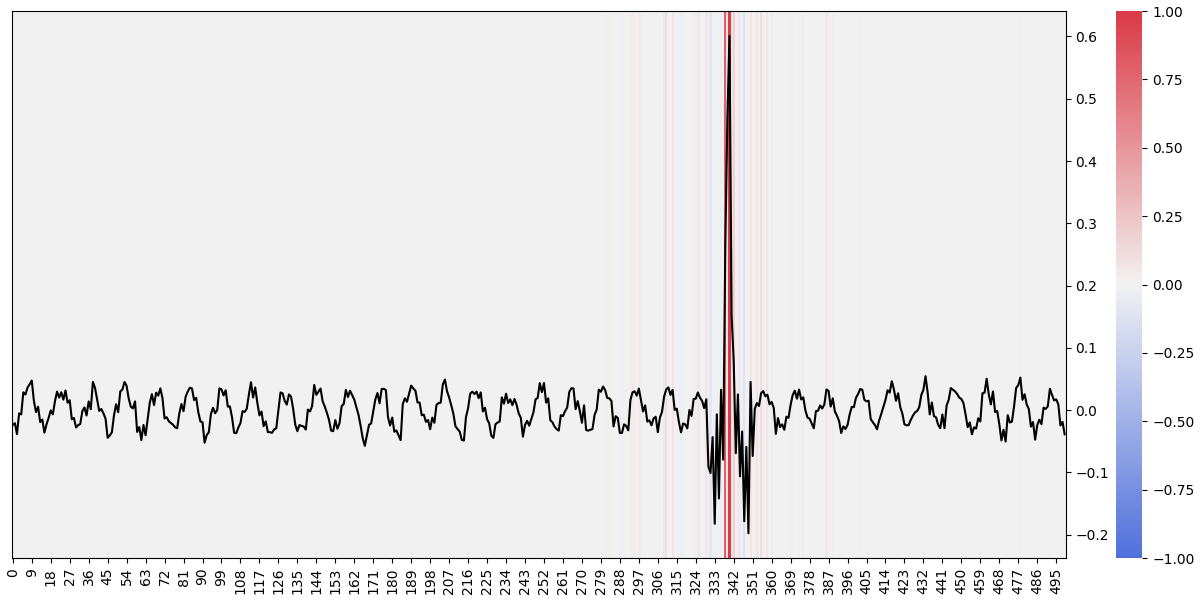}
        }
    \subfloat[Frequency]{
        \includegraphics[width=0.45\linewidth, height=0.12\textwidth]{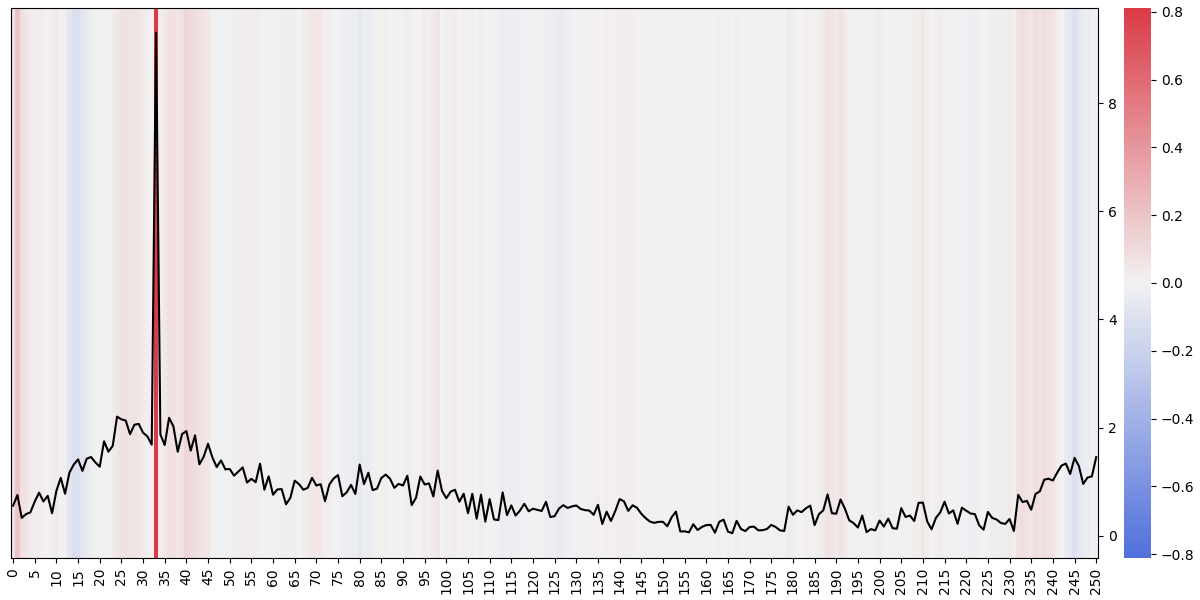}
        }
    \caption{InceptionTime trained on the synthetic dataset and the attributions are generated using DeepLIFT. An example where non-feature frequency is captured by DeepLIFT.}
    \label{fig:synthetic-InceptionTime}
\end{figure}

\begin{figure}[]
    \centering
    \subfloat[Synthetic - FCN]{%
        \includegraphics[width=0.45\linewidth, height=0.18\textwidth]{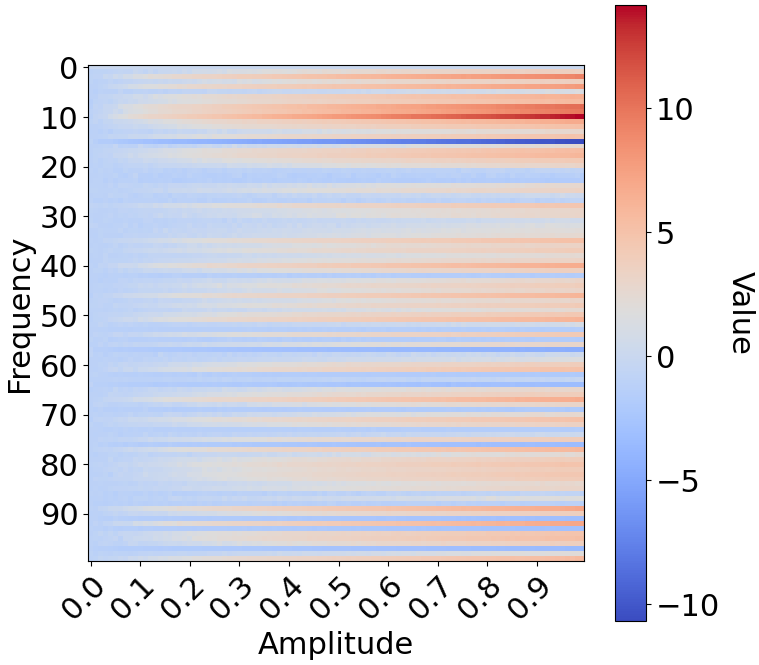}
        }
    \subfloat[Synthetic - TST]{
        \includegraphics[width=0.45\linewidth, height=0.18\textwidth]{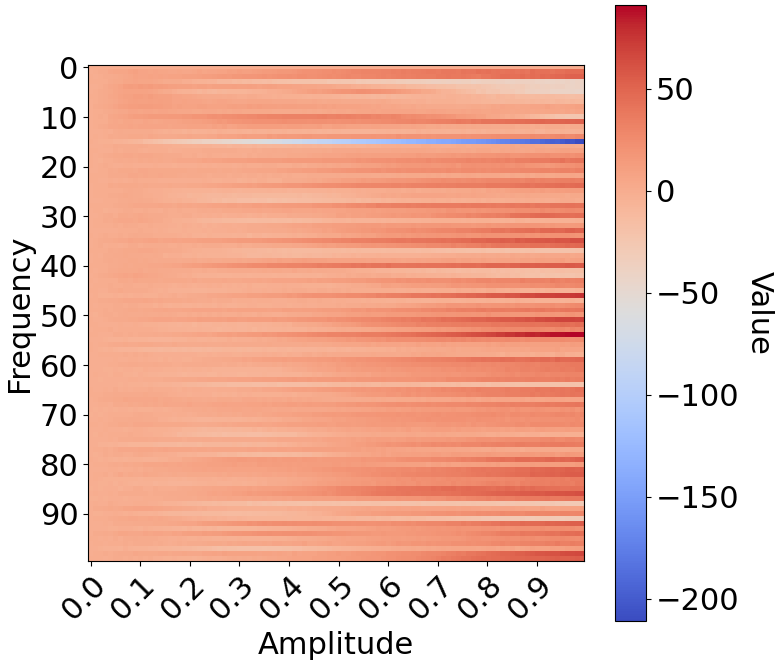}
        }
    \\
    \subfloat[Synthetic - ResNet]{%
        \includegraphics[width=0.45\linewidth, height=0.18\textwidth]{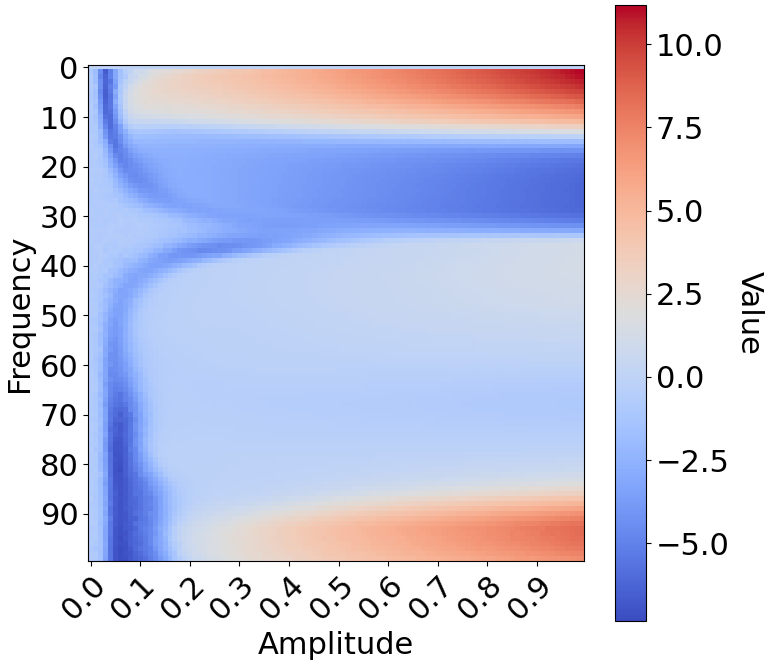}
        }
    \subfloat[Synthetic - InceptionTime]{
        \includegraphics[width=0.45\linewidth, height=0.18\textwidth]{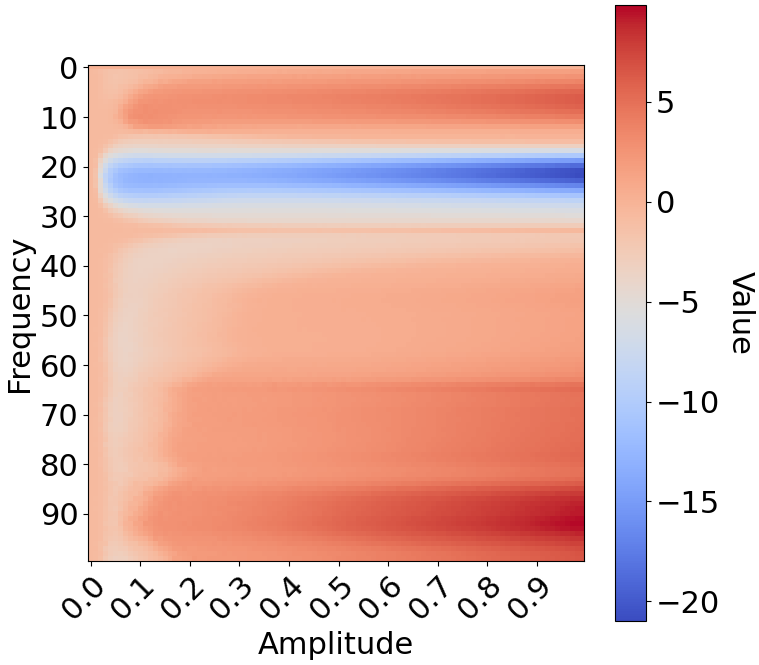}
        }
    \\
    \subfloat[FordA - TCN]{%
        \includegraphics[width=0.45\linewidth, height=0.18\textwidth]{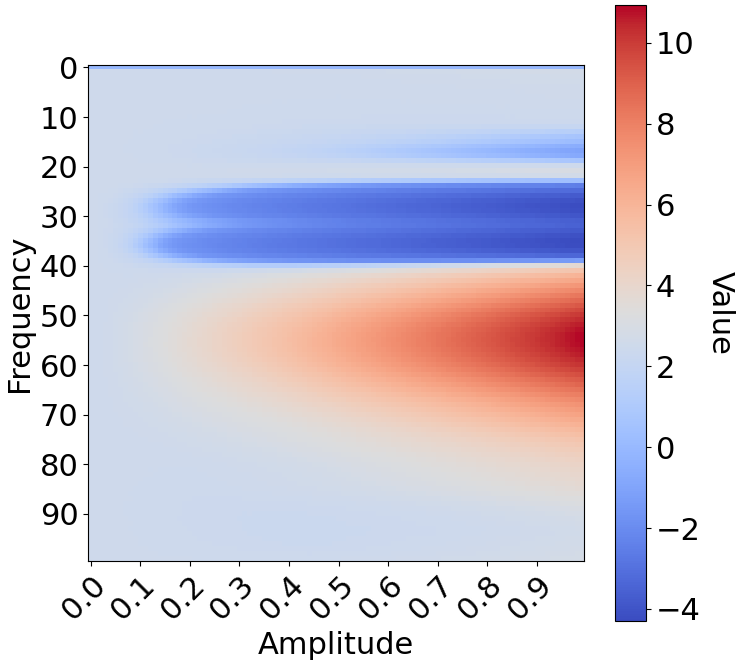}
        }
    \subfloat[MIMICPerformECG - ResNet]{
        \includegraphics[width=0.45\linewidth, height=0.18\textwidth]{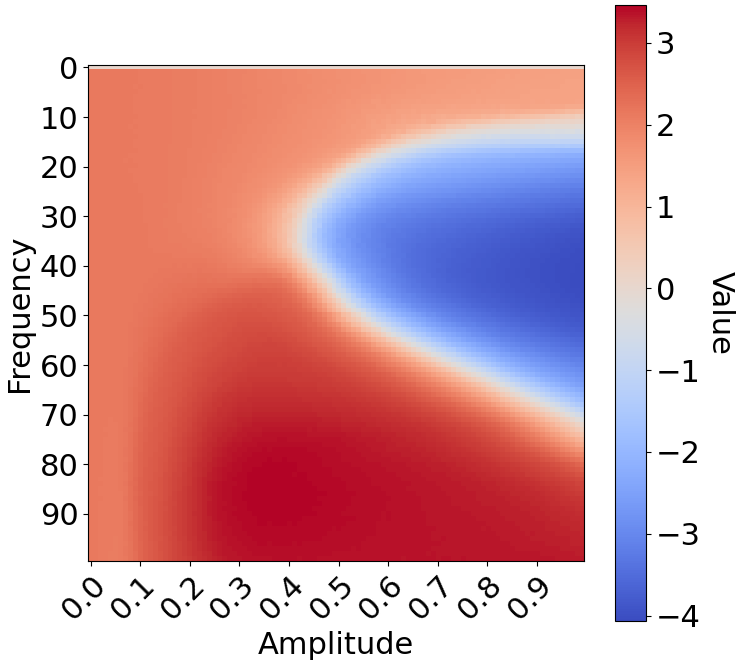}
        }
    \caption{Frequency/Amplitude response of various models trained on different datasets.}
    \label{fig:amp-fre}
\end{figure}

As shown in Table \ref{tbl-synth} for the synthetic dataset, there are cases where UP is violated while the target sample has only one feature in one domain. For example, DeepLift shows $42\%$ UP violation for class 0 samples with only time (shapelet) features. Figure \ref{fig:synthetic-InceptionTime} demonstrates such a sample where the frequency peak in the frequency domain is actually a non-feature despite being highlighted. To shed light on why, in some cases, the frequency non-feature is highlighted in some XAI methods, we plot the frequency/amplitude response of a trained model, as shown in Figure \ref{fig:amp-fre}. Here, we probe the model with a single sine wave, plus a small amount of Gaussian noise, and plot the activation of one class. We mainly focus on a binary classification for the simplicity of the plot. Here, positive values are mostly associated with the prediction of the target class, while negative values are associated with the prediction of the other class. As shown in Figure \ref{fig:amp-fre}, there is a huge discrepancy between convolution-based architectures and non-convolution-based ones. Basically, for FCN and TST, there is no blend in of adjacent frequencies when amplitude is changed. However, this is not true for convolution-based methods. The reason is that convolution filters can have essentially the same response for a peak with high-frequency/low-amplitude and a peak with low-frequency/high-amplitude. For this reason, the ground truth frequencies in the dataset generation process should not be looked at face value. Hence, frequency non-features in certain amplitude regimes can trigger one of the classes. Since the model is truly responsive to such non-features, the XAI method is correct to pick it up although it was not part of the explanation ground truth of the dataset.

\section{Discussion}
We want to emphasize that UP is only the sufficient condition to understand if the features highlighted in time versus frequency domains are not the same. Lack of UP violation makes the case inconclusive. Our current approach is conservative, which essentially gives no false positive, widely underestimating the number of samples with localized attributions in both domains. However, in practice and as in future work, we explore how one can loosen the UP bound to navigate the false positive and false negative trade-off.

Note that none of the existing XAI methods is specifically designed to localize different features in different domains if needed. LIME enforces sparsity, but it cannot reverse it if the feature is not inherently sparse in the target domain. Hence, the future work can focus on designing an XAI method specific for time series application where it can reliably localized both features if both exist in different domains and also being able avoid localizing the attribution when there is no localized feature in that domain. We propose one such method in Section \ref{sec:lime}, but further improvement are necessary.

\bibliographystyle{IEEEtran}
\bibliography{IEEEabrv,mybibfile}

\end{document}